\documentclass{article} % For LaTeX2e
\usepackage{iclr2027_conference,times}

\usepackage[utf8]{inputenc} % allow utf-8 input
\usepackage[T1]{fontenc}    % use 8-bit T1 fonts
\usepackage{hyperref}       % hyperlinks
\usepackage{url}            % simple URL typesetting
\usepackage{booktabs}       % professional-quality tables
\usepackage{amsfonts}       % blackboard math symbols
\usepackage{nicefrac}       % compact symbols for 1/2, etc.
\usepackage{silence}
\usepackage{microtype}      % microtypography
\usepackage{xcolor}         % colors

\usepackage{amssymb}            % Defines common symbols like \mathbb R
\usepackage{mathtools}          % Extends amsmath, providing common math tools
\usepackage{mathrsfs}           % Enables \mathscr
\usepackage{graphicx}           % For including images
\usepackage{subcaption}         % Allows for the use of subfigures and subcaptions
\usepackage[space]{grffile}     % For spaces in image names

\title{Gradient-Momentum Coupling: A Parameter-Space Proxy for Learning Progress}

\author{%
  Samuel Blad \\
  Örebro University\\
  \texttt{samuel.blad@oru.se} \\
  \And
  Martin Längkvist \\
  Örebro University\\
  \texttt{martin.langkvist@oru.se} \\
  \And
  Amy Loutfi \\
  Örebro University\\
  \texttt{amy.loutfi@oru.se} \\
}

\iclrfinalcopy % TODO Uncomment for camera-ready version, but NOT for submission.
\begin{document}

\maketitle
\lhead{Preprint}

\begin{abstract}
Measuring learning progress is at the core of curiosity-driven exploration, which rewards an agent for going where its model is still learning.
However, the abstract notion of learning progress is not directly measurable, and existing methods often derive it from the prediction error in the output space.
This paper proposes Gradient-Momentum Coupling (GMC), which measures how strongly a sample drives change in the parameter space, given by the normalized absolute product of its gradient with the momentum of previous gradients.
Directions of change that persist across samples accumulate in momentum, while noise cancels out.
In controlled experiments GMC allocates near uniform priority across tasks with varying levels of noise, where prediction error chases the noisiest, and orders learnable tasks by improvement speed rather than difficulty.
On four MiniGrid MultiRoom tasks, substituting GMC for prediction error inside the Intrinsic Curiosity Module (ICM) recovers exploration ICM loses to unpredictable observations.
\end{abstract}

%%%%%%%%%%%%%%%%%%%%%%%%%%%%%%%%%%%%%%%%%%%%%%%%%%%%%%%%%%%%%%%%
%% Section: Introduction and Background
%%%%%%%%%%%%%%%%%%%%%%%%%%%%%%%%%%%%%%%%%%%%%%%%%%%%%%%%%%%%%%%%

\section{Introduction and Background}\label{sec:intro}

A central challenge in reinforcement learning is efficient exploration, particularly in environments with sparse or deceptive rewards \citep{sutton1998introduction}. 
Intrinsic motivation signals address this by providing internal rewards that guide sample selection, curriculum design, and prioritization \citep{schmidhuber2010formal, oudeyer2007intrinsic}. 
The goal is to identify where the agent's model is still making progress, a goal we share with the prior literature.
What counts as progress, however, is not settled.

\citet{schmidhuber1991possibility} proposed rewarding an agent for the reduction in prediction loss that an observation produces.
In the approaches that follow, a learned dynamics model predicts some aspect of the environment (e.g., the next observation), and the intrinsic reward is derived from how well it does so.
Intelligent Adaptive Curiosity \citep{oudeyer2007intrinsic} estimates that reduction empirically within regions of the sensorimotor space, while the Intrinsic Curiosity Module (ICM) \citep{pathak2017curiosity} and Random Network Distillation (RND) \citep{burda2019exploration} use the prediction error itself, under the assumption that large errors mark situations worth prioritizing.
These differ in what is predicted and in whether the loss or its change is used, but all of them read progress off the model's output.

\paragraph{Failure Modes of Loss-Based Curiosity.}
Two critical issues undermine standard loss-based curiosity approaches. 
First, the \textit{noisy-TV problem} \citep{burda2018large}: when prediction errors arise from aleatoric noise (irreducible randomness inherent in the environment, i.e.\ television static), they indicate uninformative surprises, not learning opportunities. 
An agent optimizing for prediction error will perpetually be drawn to noisy transitions, even after all learnable structure has been captured \citep{burda2018large}. 
Even methods that modify the representation \citep{pathak2017curiosity} or prediction target \citep{burda2019exploration} still use instantaneous prediction error as the intrinsic signal, a proxy that remains susceptible to confounding noise with learnable structure.

Second, \textit{the loss is an indirect view of what a model has learned}, and the two can come apart. 
Training can plateau followed by abrupt improvement \citep{power2022grokking}, and transient increases in loss can precede better performance.
A signal defined as change in loss reports no progress across such a plateau and negative progress during such an increase, which would say that the model stopped learning and then unlearned, rather than that the loss is an incomplete view of its state.
The decoupling may also run the other way, since a model can build structure of which only some part ever helps the task at hand.
In reinforcement learning the distance between the two is larger still, because the agent is judged on return rather than on fit, and because the data the model learns from is produced by the agent's own exploration rather than given in advance.

To address these issues, we propose to measure change in the model's parameters directly, rather than change in loss. 
Momentum, the exponentially-weighted average of recent gradients, accumulates the directions that persist across samples and cancels those that do not, filtering random noise (which points in inconsistent directions across samples) and oscillatory dynamics (which cancel over time). 
A hard but learnable task may initially produce weakly aligned gradients, and as a useful representation forms its persistent component grows and accumulates in momentum.
Pure noise, by contrast, has no sustained direction to accumulate once its predictable statistics have been fitted.
GMC therefore prioritizes samples that drive the directions the model is already moving in, and defers weakly aligned ones, which can be confounded with noise, until the representation is more mature. 
The momentum vector over all model parameters reveals which parameters are currently undergoing the most change, and comparing a sample's gradient with this per-parameter activity signal enables prioritization by the proposed proxy.

In this paper, we denote \emph{task performance} as the quantity an agent is ultimately judged on, the predictive model's loss or the policy's return, and we reserve \emph{learning progress} for change in what the model has internalized, which may not always be directly reflected in the loss.
Learning progress in this sense is an abstract target rather than a defined quantity, and prediction error, its change over time, and GMC are all proxies for it.
Persistent parameter change is \emph{necessary} for a model to train at all, but like a falling loss it is not \emph{sufficient} for task performance to improve, since some of what a model internalizes may never turn out to be useful for the task at hand.

\paragraph{Relation to Prior Work.}
Our approach builds on learning-progress intrinsic motivation \citep{schmidhuber2010formal, oudeyer2007intrinsic}, but shifts the measurement from external performance metrics to an internal quantity, parameter change as observed through gradient-momentum interaction. 
Prior formulations operationalize learning progress through observable changes in predictive performance, such as Schmidhuber's \emph{prediction gain} \citep{schmidhuber1991possibility, schmidhuber2010formal} and Intelligent Adaptive Curiosity \citep{oudeyer2007intrinsic}, which takes the empirical derivative of prediction error within regions of the sensorimotor space.
Learning Progress Monitoring (LPM) \citep{hou2026beyond} also targets noise-robust exploration, and does so in output space by training a second model of how the prediction error evolves, which refines the estimator of a loss-difference quantity rather than the quantity that is read.

This is complementary to curiosity architectures that determine \emph{what} to predict: ICM \citep{pathak2017curiosity} uses inverse dynamics to filter uncontrollable features, RND \citep{burda2019exploration} predicts fixed random targets for novelty, and Variational Information Maximizing Exploration (VIME) \citep{houthooft2016vime} maximizes information gain. 
Recent work on multi-step inverse dynamics \citep{efroni2021provably, islam2023principled} extends single-step filtering to provably remove exogenous distractors, improving the \emph{representation} on which curiosity is measured.
GMC scores learning on whichever representation these methods produce, and could replace the prediction error used within them.

Curriculum learning \citep{bengio2009curriculum, graves2017automated} designs explicit training schedules that progress from easy to hard, whereas GMC proposes a signal from which such prioritization can emerge directly, without manual scheduling.

Prior work has also used gradient information for sample prioritization. 
Gradient-norm methods \citep{katharopoulos2018not, zhao2015stochastic} weight samples by $\|\nabla_\theta \mathcal{L}\|$ to reduce variance in SGD updates, and CurES \citep{zeng2025cures} selects samples whose gradients align with a held-out validation set, both targeting \emph{optimization efficiency} in supervised settings with reliable labels and a fixed dataset.
GMC instead targets online selection under aleatoric noise, since momentum-based filtering provides robustness to noise and the signal can be computed without reference to a validation set. 

Closest within RL, \citet{linke2020} study weight change $\lVert\Delta w\rVert$ as an intrinsic reward for tabular parallel general value functions.
This rewards the magnitude of change but not whether successive changes share a direction, making it closest to the gradient-magnitude baseline we call NormAll (Section~\ref{sec:controlled}), which is degraded by aleatoric noise in our controlled experiments, whereas GMC couples the gradient to the per-parameter momentum and operates over deep dynamics models.
See Appendix~\ref{sec:app_related} for an extended discussion of related work.

%%%%%%%%%%%%%%%%%%%%%%%%%%%%%%%%%%%%%%%%%%%%%%%%%%%%%%%%%%%%%%%%
%% Section: Method
%%%%%%%%%%%%%%%%%%%%%%%%%%%%%%%%%%%%%%%%%%%%%%%%%%%%%%%%%%%%%%%%

\section{Method}\label{sec:method}

Section~\ref{sec:intro} motivated parameter dynamics as a proxy for learning progress, using momentum as a noise-filtered summary of ongoing change.
The goal is to quantify how strongly a single sample's gradient contributes to ongoing parameter change.
Consider the momentum $\mathbf{m} \in \mathbb{R}^d$ summarizing the current direction and magnitude of change across all $d$ model parameters.
A single sample's dynamics loss gradient $\mathbf{g} = \nabla_\theta \mathcal{L}_{dyn}(x) \in \mathbb{R}^d$ enters $\mathbf{m}$ through the exponential moving average, $\mathbf{m} \leftarrow \beta \mathbf{m} + (1-\beta)\mathbf{g}$, so the increment it induces, $\Delta_\mathbf{m} = (1-\beta)\left(\mathbf{g} - \mathbf{m}\right)$, is small relative to $\|\mathbf{m}\|$.
A first-order expansion of the induced change in momentum magnitude, $\|\mathbf{m} + \Delta_\mathbf{m}\| - \|\mathbf{m}\|$, splits into a term that is identical for every sample and a term that is not.
Appendix~\ref{sec:app_taylor} derives the sample-dependent part, the \emph{relative impact} of a sample on the momentum magnitude, up to a positive factor shared across samples, as:
\begin{equation}
\rho(x) \;=\; \frac{\mathbf{m}^{\top} \mathbf{g}}{\|\mathbf{m}\|}
\label{eq:first_order}
\end{equation}
This ranks samples by how strongly they accelerate or decelerate the ongoing parameter changes.
Both reinforcing and opposing the current direction indicate the sample is \emph{relevant} to actively-learned parameters, so the absolute value should be applied.
Without this, the method would preferentially sample confirmatory examples.

\paragraph{Gradient-Momentum Coupling (GMC).}
Equation~\ref{eq:first_order} measures the sample's influence on momentum as a single vector.
We instead measure influence per parameter, applying the absolute value elementwise so that parameters changing in different directions contribute independently.
For normalization across parameters, we replace $\|\mathbf{m}\|$ with a per-parameter scale
\begin{equation}
s_i = \max\!\left(\hat v_i,\,\alpha^2 \hat q_i\right),
\label{eq:gmc_scale}
\end{equation}
where $\hat v_i$ and $\hat q_i$ are the exponentially-weighted means of the squared gradient and squared parameter, respectively, and $\alpha$ is a constant.
Different parameters can have gradient magnitudes that differ by orders of magnitude, so a global norm would let high-magnitude parameters dominate the signal.
The maximum retains the gradient-based normalization when it is informative, while preventing a very small second moment from amplifying the score relative to the current parameter scale.
Appendix~\ref{sec:app_ablation} compares design choices empirically.
Formally, GMC is defined as:
\begin{equation}
r_{\text{gm}}(x) = \frac{1}{\sqrt d}\sum_i \left|\frac{\nabla_{\theta_i} \mathcal{L}_{dyn}(x) \cdot \hat m_i }{s_i}\right|
\label{eq:r_gm}
\end{equation}
Here $x$ is the current sample, $i$ indexes all $d$ model parameters, $\nabla_{\theta_i} \mathcal{L}_{dyn}(x)$ is the gradient of the dynamics loss with respect to parameter $\theta_i$, and $\hat m_i$ is the gradient momentum.
% The numerator $\nabla_{\theta_i}\mathcal{L}_{dyn}(x)\hat m_i$ and both terms in $s_i$ have units of gradient squared (with $\alpha$ carrying the corresponding relative gradient scale), so their ratio is dimensionless. 
The score uses statistics accumulated through the previous batch and updates $m$, $v$, and $q$ only after scoring the current samples. Thus a sample is compared against prior learning dynamics rather than statistics containing itself.
$r_{\text{gm}}(x)$ adds $O(d)$ storage and elementwise work for $d$ scored parameters, the same asymptotic order as the backward pass (see Appendix~\ref{sec:app_implementation}).

We compute GMC on a reward-independent dynamics model rather than on the policy or value parameters.
Scoring samples by the policy's own gradients would make the intrinsic reward depend on the updates it induces, and before any extrinsic reward is found those gradients carry little stable signal.
A predictive dynamics objective is dense and grounded in observed transitions, and it lets GMC replace the prediction-error term used by architectures such as ICM.

%%%%%%%%%%%%%%%%%%%%%%%%%%%%%%%%%%%%%%%%%%%%%%%%%%%%%%%%%%%%%%%%
%% Section: Experiments
%%%%%%%%%%%%%%%%%%%%%%%%%%%%%%%%%%%%%%%%%%%%%%%%%%%%%%%%%%%%%%%%

\section{Experiments}\label{sec:experiments}

We pose three questions: (1) Does GMC produce meaningfully different task prioritization than loss-based signals? (2) Can GMC distinguish learnable structure from aleatoric noise? (3) Does GMC improve exploration when it replaces prediction error inside a full RL pipeline?

To answer (1) and (2), we design controlled adaptive-sampling experiments where task difficulty and noise can be precisely manipulated, isolating the learning-progress signal from temporal credit assignment and policy optimization.
To answer (3), we compute ICM's intrinsic reward with GMC on a family of sparse-reward MiniGrid navigation tasks.

\subsection{Signal Evaluation under Controlled Difficulty and Noise}\label{sec:controlled}

\subsubsection{Environment}

This reduced environment provides no extrinsic reward. 
The agent's sole objective is to prioritize tasks according to the learning-progress proxy supplied as its intrinsic signal.
This mimics the situation in sparse-reward RL where no reward has yet been discovered and exploration must be guided entirely by intrinsic motivation.
By observing which samples each signal prioritizes, we can infer its implicit answer to what counts as progress, and how it tackles different levels of noise.

The agent consists of an actor and an image classifier, which we call the dynamics model by analogy with RL.
Every episode is a single step, as in a multi-armed bandit setting \citep{sutton1998introduction}.
During the episode: the actor selects an action, without receiving any environment or task state; the environment returns an image together with a \emph{target} drawn from the 10 classes; the dynamics model predicts that target and the episode ends.
This is analogous to an RL agent choosing which part of an environment to visit, each of which presents different learning opportunities, but without state or temporal credit assignment.
The dynamics loss $\mathcal{L}_{dyn}$ is the cross-entropy between the dynamics model's 10-way prediction and the target supplied by the environment.
To emulate observations, we sample images from MNIST \citep{lecun1998gradient} and CIFAR-10 \citep{krizhevsky2009learning}.
Targets are reassigned within groups, as described below, so a target is in general not the image's original dataset label.
Each of the 10 dataset classes is assigned to one of four groups: $G_0=\{0\}$, $G_1=\{1,2\}$, $G_2=\{3,4,5\}$, and $G_3=\{6,7,8,9\}$.
We test two conditions that differ in what an action selects and in how the target is drawn, giving controlled variation in task difficulty and in irreducible noise, targeting questions (1) and (2) respectively:

\paragraph{Condition 1 - Curriculum:} Before training, each image is permanently assigned a target drawn uniformly from the classes of its own group (e.g., a cat image becomes ``dog'') and keeps it for the whole run.
The target is therefore a deterministic and fully learnable, but arbitrary, function of the image.
The 10 reassigned classes are the actions: selecting class $c$ returns an image whose fixed target is $c$, so here the target coincides with the chosen action.
The groups become four tasks of differing intrinsic difficulty: the single class of $G_0$ is untouched, whereas each class of $G_3$ pools images from four visually distinct source classes and must essentially be memorized.
We report results per group by aggregating the classes belonging to each group.

\paragraph{Condition 2 - Noise:} The four groups are the actions, and the target is redrawn uniformly from the selected group's classes \textit{each time} an image is drawn, so it is unpredictable from the image and the action fixes only the distribution the target comes from.
Groups $G_0$--$G_3$ therefore carry 0\%, 50\%, 67\%, and 75\% irreducible label noise respectively (larger groups offer more equally valid targets).

\subsubsection{Compared Signals}

We compare six methods for computing intrinsic rewards from the dynamics model:

\begin{itemize}
\setlength{\itemsep}{1pt}\setlength{\parskip}{0pt}\setlength{\parsep}{0pt}
    \item \textbf{Uniform}: No intrinsic reward $r_{\text{int}}=0$; baseline with random sampling across groups.
    \item \textbf{Curiosity}: Prediction loss of the classifier $r_{\text{int}} = \mathcal{L}_{dyn}$; canonical curiosity baseline using loss as proxy for learning.
    \item \textbf{Gradient-Momentum Coupling (GMC)}: $r_{\text{int}} = d^{-1/2}\sum_i |\nabla_{\theta_i} \mathcal{L}_{dyn} \cdot \hat m_i / \max(\hat v_i,\alpha^2\hat q_i)|$; our proposed method.
    \item \textbf{NormOut}: Squared $L_2$ norm of the gradient of the dynamics loss with respect to the model's output, $r_{\text{int}} = \|\nabla_{z}\mathcal{L}_{dyn}\|_2^2$, where $z$ denotes the 10 output logits; like Curiosity it is measured in output space, but it looks at gradient magnitude rather than loss itself.
    \item \textbf{NormAll}: Squared $L_2$ norm of the gradient of the same loss with respect to all $d$ model parameters, $r_{\text{int}} = \|\nabla_\theta\mathcal{L}_{dyn}\|_2^2$; takes the same gradient-magnitude idea into parameter space, and is similar to GMC but without the momentum weighting and per-parameter normalization.
    \item \textbf{DeltaLoss}: $r_{\text{int}}=|\bar{\mathcal L}_{g,t-1}-\bar{\mathcal L}_{g,t-2}| / (N_{g,t-1}+N_{g,t-2})$, a numerical approximation of the derivative of group $g$'s loss with respect to the number of samples that group has received%. Here $\bar{\mathcal L}_{g,t}$ is the mean loss on group $g$ during epoch $t$ and $N_{g,t}$ is the number of samples it received. Samples rather than epochs are the natural denominator, since adaptive allocation gives each group a different number of samples per epoch. This 
    ; tests direct learning-progress estimation from observed loss changes.
\end{itemize}

The agent, or selection policy, is trained using policy gradient with entropy regularization \citep{sutton1998introduction}. 
Because the random policy input is independent of the environment, the actor learns a selection distribution over groups or classes.
The policy loss is therefore:
\begin{equation}
\mathcal{L}_{\text{policy}} = -\mathbb{E}_\pi[\log \pi(a) \cdot r_{\text{int}}] - \lambda_{\text{ent}} H(\pi)
\end{equation}
Here $r_{\text{int}}$ is the intrinsic signal from the corresponding method, $\pi(a)$ is the probability assigned to the selected action $a$ by the actor, and $\lambda_{\text{ent}}$ weights the entropy term $H(\pi)=-\sum_{i \in \text{all actions}}\pi(a_i)\log(\pi(a_i))$ to prevent collapse to a single group.
$\lambda_{\text{ent}}$ is shared across all methods.
Because the policy-gradient term scales with the magnitude of $r_{\text{int}}$, every learned signal is passed through the same automatic scale calibration, so that no method is favored or penalized against the entropy term by its raw scale alone (Appendix~\ref{sec:app_implementation}).
Increasing $\lambda_{\text{ent}}$ moves every method toward uniform allocation, while decreasing it sharpens allocations toward collapse.

At the end of each epoch, we compute a \emph{Balanced Evaluation Loss}: the average loss over an equal number of samples from each class or group.
In the Curriculum condition, evaluation uses the same fixed image-label assignments as training and therefore measures a memorization fit.
In the Noise condition, labels are re-randomized per sample, leaving an irreducible floor.
To visualize task prioritization, we report ``Training Allocation / Uniform'': the mean per-class visitation within each group during a training epoch, divided by the visitation each class would receive under uniform sampling.
All runs in this section use 20 random seeds with different dataset shuffles and network initializations.
Shaded regions show 95\% confidence intervals (CI) following \citet{patterson2024empirical}. 
These are computed as $\bar{x} \pm t_{0.975,\, n-1} \cdot \sigma / \sqrt{n}$ where $\sigma$ is the sample standard deviation and $n$ the number of seeds. 
Two variants of each experiment were run, one for the MNIST dataset and one for CIFAR-10. 
All reports are computed separately for each dataset. 
Further implementation details are available in Appendix~\ref{sec:app_implementation}.

\subsubsection{Results: Curriculum Condition}

\begin{figure}[ht]
    \centering
    \includegraphics[width=0.95\linewidth]{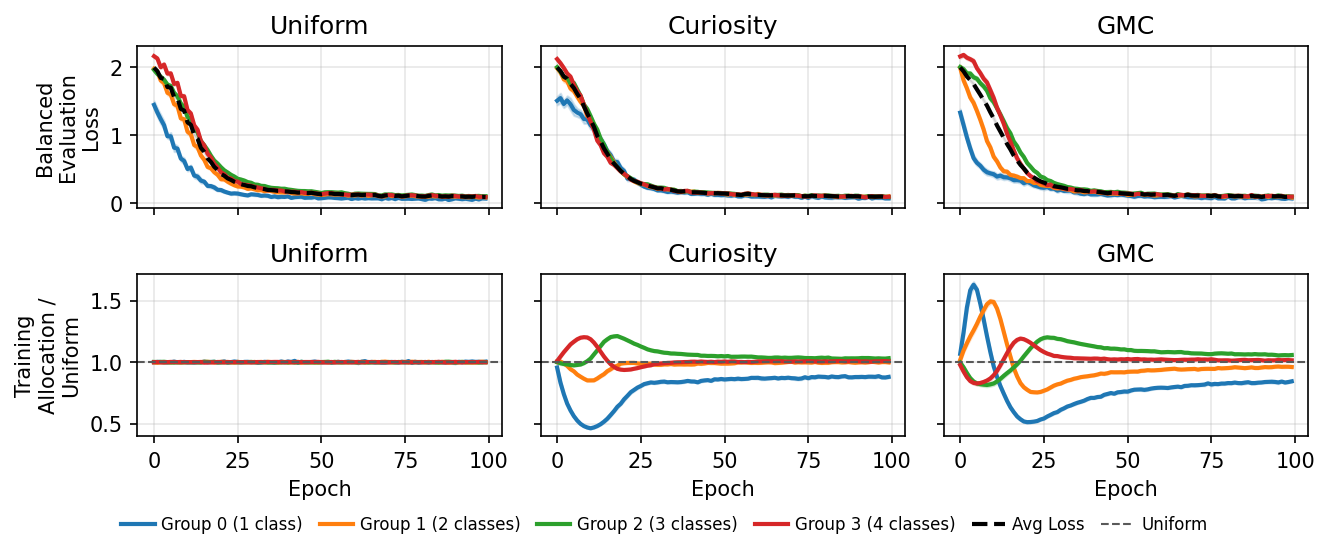}
    \caption{CIFAR-10 Curriculum Condition. \textbf{Top:} Per-group Balanced Evaluation Loss. \textbf{Bottom:} Mean per-class training allocation within each group. Arrows mark each group's visitation peak.}
    \label{fig:curriculum_cifar}
\end{figure}

This experimental condition creates tasks of varied difficulty by scrambling the labels once, over varying group sizes. 
A zero loss is theoretically achievable through memorization, but within the model's structural limitations and training time, this is not achieved. 
This section compares how Uniform, Curiosity and GMC vary in the prioritization of tasks with respect to difficulty.

Figure~\ref{fig:curriculum_cifar} reveals each method's implicit learning-progress measure through its sampling allocation. 
Uniform has no prioritization, and it reveals some differences in difficulty between the groups.
Group size controls how many labels can be assigned, but it does not determine the empirical difficulty by itself, because the particular classes placed in each group also matter.
Notably, the four-class $G_3$ starts with the highest loss of all groups under Uniform sampling, yet its curve crosses below the three-class $G_2$ partway through training.
Curiosity shifts focus based on absolute loss magnitude, which can be observed in how the losses become similar for all groups, the loss curves contract.
GMC produces an almost opposite pattern, and its loss curves spread out instead.

The difference in prioritization is clearest in the training allocation, where each group's visitation peaks at a different time.
Curiosity peaks in the order $G_3 \rightarrow G_2 \rightarrow G_1 \rightarrow G_0$, which in this case coincides with the ordering from highest initial loss to lowest.
GMC peaks in the order $G_0 \rightarrow G_1 \rightarrow G_3 \rightarrow G_2$, which appears to favor the groups whose loss is falling fastest at the time.
We see that GMC visits $G_2$ last even though it sits below $G_3$ for the first part of training under Uniform sampling.
For $G_3$ to catch up and cross below $G_2$, its loss must have been descending more steeply over that interval, and that is where GMC directs its budget.
The pattern is one of taking the quick wins first and deferring the groups whose loss is moving slowly, even when that means postponing progress on a group that is still far from converged.
GMC's prioritization of $G_0$ is visible in the early loss curves, where $G_0$ falls faster than under any other method, indicating that the early allocation is spent productively on this group.
In contrast, for Curiosity $G_0$ visitation drops rapidly at the start since it focuses on groups with higher absolute loss.
The MNIST (Appendix~\ref{sec:app_results}) results show more pronounced effects due to larger difficulty differences between groups.

To test whether this ordering reflects the particular classes placed in each group, Appendix~\ref{sec:app_repartition} repeats the experiment on two additional partitions with randomly drawn class-to-group assignments and the same group sizes, which produce visibly different difficulty orders among the groups, and GMC's visitation order changes with them.
GMC does not converge faster than Uniform or Curiosity in this condition (Figure~\ref{fig:all_conditions}), so in this simple setting the signal produces a difference in prioritization rather than a convergence advantage.

\subsubsection{Results: Noise Condition}

\begin{figure}[ht]
    \centering
    \includegraphics[width=0.95\linewidth]{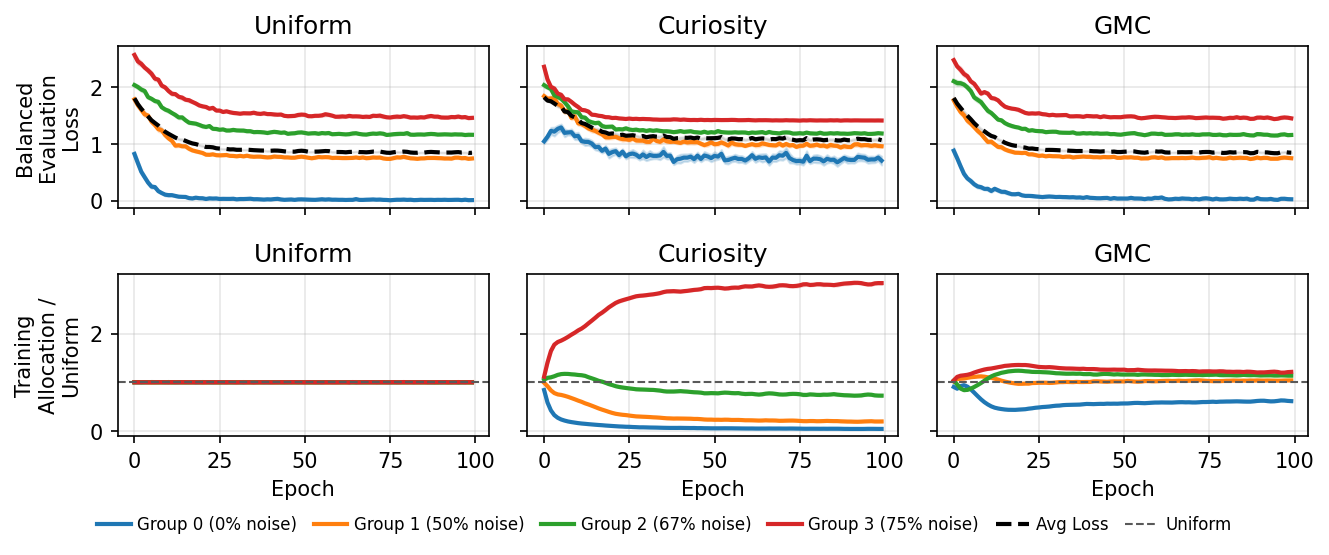}
    \caption{CIFAR-10 Noise Condition. \textbf{Top:} Per-group Balanced Evaluation Loss under uniform sampling (0\%--75\% label noise). \textbf{Bottom:} Mean per-class training allocation within each group.}
    \label{fig:noise_cifar}
\end{figure}

This experimental condition creates tasks with varied amount of aleatoric noise by scrambling the labels each time, over varying group sizes. 
A zero loss is now impossible, and methods distracted by noise are expected to fail. 
This section compares the noise robustness of Uniform, Curiosity and GMC.

As Figure~\ref{fig:noise_cifar} shows, Curiosity dramatically over-allocates to the noisiest group, whereas GMC maintains near-uniform allocation across noise levels. 
Furthermore, GMC shows a slight bias against the no-noise group after saturation. 
This could be expected from the numerical properties of its expression. When losses plateau the momentum becomes small $\mathbf{m} \approx \mathbf{0}$, however the gradients are only small for the no-noise group $|\nabla_{\theta_i} \mathcal{L}_0| \approx 0$. 
In the no-noise case, GMC multiplies two near zero values, whereas in the other groups only one value is near zero.
The MNIST variant (Appendix~\ref{sec:app_results}) shows similar patterns.

\subsubsection{Overall Convergence Across Methods}\label{sec:summary_metrics}

\begin{table}[t]
    \centering
    \small
    \setlength{\tabcolsep}{4pt}
    \begin{tabular}{lcccccc}
        \toprule
        Condition & Uniform & Curiosity & GMC & NormOut & NormAll & DeltaLoss \\
        \midrule
        Curriculum & $1.00 \pm 0.01$ & $0.99 \pm 0.00$ & $1.01 \pm 0.00$ & $1.00 \pm 0.00$ & $1.00 \pm 0.00$ & $\mathit{1.14 \pm 0.01}$ \\
        Noise & $1.00 \pm 0.00$ & $\mathit{1.23 \pm 0.00}$ & $0.99 \pm 0.00$ & $\mathit{1.34 \pm 0.00}$ & $\mathit{1.07 \pm 0.00}$ & $1.02 \pm 0.00$ \\
        \bottomrule
    \end{tabular}
    \caption{Normalized area under the overall Balanced Evaluation Loss curve on CIFAR-10, scaled so that Uniform is $1.00$ in each condition. This table reports failures in italics.}
    \label{tab:auc_cifar}
\end{table}

The per-group figures above cover Uniform, Curiosity and GMC.
Table~\ref{tab:auc_cifar} summarizes all six methods on CIFAR-10 by the area under the overall Balanced Evaluation Loss curve.
In the noise condition Curiosity and NormOut are the two methods clearly above the Uniform level, both are driven by the magnitude of an output-space error and spend their budget on the noisiest groups.
NormAll sits between them and Uniform, so moving the same gradient-magnitude idea from output space into parameter space reduces the damage.
GMC and DeltaLoss stay at the Uniform level, and the loss curves in Appendix~\ref{sec:app_results} show that they and Uniform reach the same final loss, with DeltaLoss slower to get there.
In the curriculum condition all six methods end at the same loss, and only DeltaLoss converges visibly more slowly.

This disqualifies NormOut and NormAll as noise-robust signals, and leaves DeltaLoss as the one baseline that filters noise as GMC does.
However, DeltaLoss estimates progress from outside the model by differencing per-task loss averages over completed epochs, which requires those averages to be tracked per task and makes the estimate visibly unstable, as the oscillations in its allocation show. 
LPM \citep{hou2026beyond} takes that approach further and addresses the instability by fitting a separate model of how the prediction error evolves, which makes it a trained component with its own dynamics rather than a quantity that can be substituted for $r_{\text{int}}$ in place, as every signal compared here is.
DeltaLoss reaches the same prioritization order as GMC here, but GMC arrives at that order by measuring parameter dynamics directly, without externally tracking an estimate of rate of change.
Per-group losses and allocations for NormOut, NormAll and DeltaLoss in all four conditions are given in Appendix~\ref{sec:app_results}.

\subsection{MultiRoom Benchmark: GMC as a Filter for ICM}\label{sec:minigrid}

Having analyzed GMC's properties in isolation, we next integrate it into a full RL pipeline on a standard sparse-reward benchmark, addressing question~(3).
GMC and ICM \citep{pathak2017curiosity} address different parts of the problem, as discussed in Section~\ref{sec:intro} and Appendix~\ref{sec:app_related}. 
ICM determines which features the dynamics model represents, while GMC determines how a sample is scored on those features.
ICM's inverse-dynamics objective does not remove unpredictability from the signal, it relocates it into the learned feature space.
MultiRoom lets us test a scenario where prediction error degrades as an intrinsic signal wherever the feature space retains variation the forward model cannot resolve.
We keep the ICM architecture and its training objective unchanged and replace only the quantity used as the intrinsic reward, which isolates whether scoring the same features with GMC avoids that degradation. 

\subsubsection{Setup}

\begin{figure}[ht]
    \centering
    \includegraphics[width=0.90\linewidth]{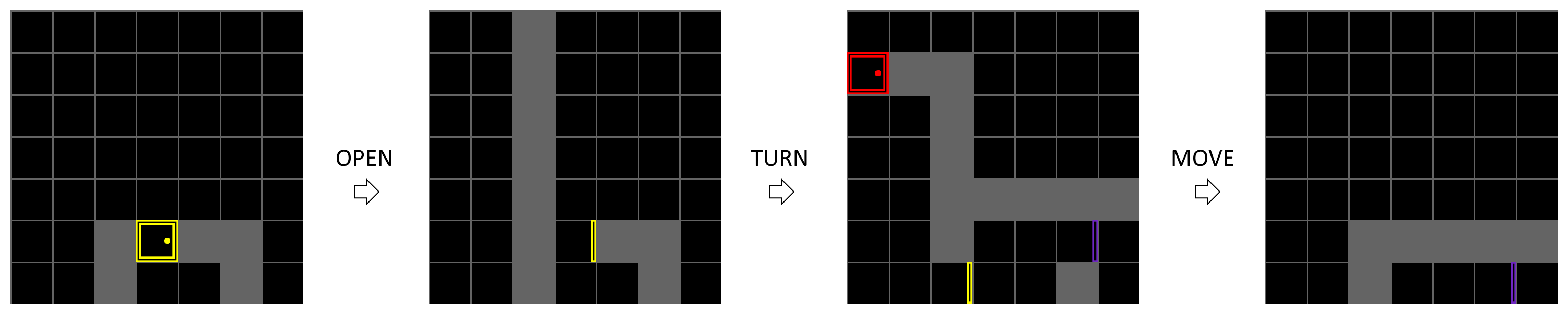}
    \caption{Four consecutive observations from a MultiRoom episode, with the action taken between them shown above the arrows. Opening a door reveals part of the next room, turning right and moving forward hides it again. The view is egocentric and partial, so what a rotation brings into view is not contained in the observation that precedes it.}
    \label{fig:multiroom_obs}
\end{figure}

We evaluate on the MiniGrid MultiRoom family \citep{chevalier2023minigrid}, in which the agent must pass through a chain of rooms connected by closed doors and reach a goal tile in the final room.
A single sparse reward is given on reaching the goal, and no other reward is provided.
Observations are partial and egocentric, so the agent sees a small view cone in front of itself and receives no information about what lies behind it.
The relevant actions are turn left, turn right, move forward, and toggle, for opening doors.
Figure~\ref{fig:multiroom_obs} shows a short sequence of observations and the actions that produced them.
We run four instances of increasing sizes (2, 3, 4, and 6), with the step budget scaled to the difficulty of each.
Each configuration is run for 50 random seeds, and shaded regions show 95\% confidence intervals computed as in Section~\ref{sec:controlled}.

\paragraph{Compared Methods.}
We compare three methods, all built on Proximal Policy Optimization (PPO) \citep{schulman2017proximal}.
\textbf{PPO} is vanilla PPO with no intrinsic reward.
\textbf{ICM} is PPO augmented with ICM \citep{pathak2017curiosity}, where a learned feature encoder, inverse dynamics model, and forward dynamics model in the learned feature space provide intrinsic reward as the forward model's prediction error.
\textbf{ICM+GMC} retains the full ICM architecture and trains it with the standard ICM loss, but computes the intrinsic reward as GMC (Equation~\ref{eq:r_gm}) on the ICM forward model rather than as its prediction error.

\subsubsection{Results}

\begin{figure}[ht]
    \centering
    \includegraphics[width=0.98\linewidth]{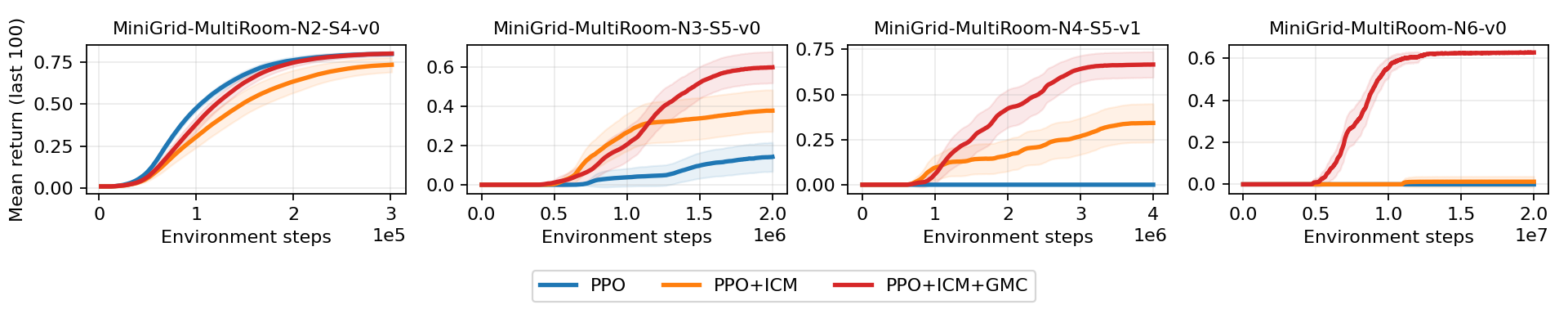}
    \caption{MiniGrid MultiRoom mean episodic return. Each panel is one environment, ordered by increasing size and step budget. The advantage of ICM+GMC over ICM grows as the task gets harder.}
    \label{fig:multiroom_return}
\end{figure}

The advantage of ICM+GMC over ICM grows with environment size (Figure~\ref{fig:multiroom_return}).
In the smallest environment exploration is not the bottleneck, and the fact that ICM ends just below PPO there indicates that its intrinsic reward already biases the policy away from the extrinsic optimum when no exploration help is needed, which computing the same reward with GMC avoids.
From three rooms onward the two intrinsic methods separate in the order ICM+GMC, ICM, and the gap between them persists once PPO stops reaching the goal altogether, so the sparse reward is findable without intrinsic motivation only in the smaller instances.
By six rooms ICM's advantage over PPO has essentially vanished although the same intrinsic reward was clearly useful at three and four rooms, whereas ICM+GMC still solves the task, so what degrades with scale is the quality of the intrinsic signal rather than the usefulness of having one.

The MultiRoom observation is egocentric and partial, and the turning actions rotate the view cone.
What a turn reveals is not contained in the observation preceding it (Figure~\ref{fig:multiroom_obs}), so the forward model cannot predict it from the current state and action however long it trains, and from the model's point of view the residual is aleatoric.
ICM's inverse-dynamics objective does not suppress this and to some extent encourages it, since rotations are among the actions most easily identified from a pair of consecutive observations and keeping the visible layout in the encoding is what makes them identifiable.
The features that make the inverse model accurate are therefore the same ones whose forward prediction stays unreliable.
Its consequence is that ICM's prediction error keeps rewarding rotation, so the larger the environment the more of the budget is spent turning rather than crossing rooms.
GMC scores the same features but is robust to unpredictable variation, for the reasons described in Section~\ref{sec:controlled}, so unlike ICM it is not drawn to the rotation residual.

Appendix~\ref{sec:app_multiroom} reports the behavioral quantities that this account predicts.
In all four environments ICM assigns a persistently larger share of its intrinsic reward to rotation than ICM+GMC does, and correspondingly turns more and moves forward less, with the difference widening as the environments grow.

\section{Discussion and Conclusion}\label{sec:discussion}

The controlled experiments reveal noise robustness and distinct prioritization behavior, while the MultiRoom study shows that substituting GMC for prediction error inside ICM restores exploration that ICM loses on the MultiRoom family of tasks.

\paragraph{The Value of Depth.}
NormOut and NormAll apply the same gradient-magnitude idea in output space and in parameter space respectively, so comparing them isolates what measuring in parameter space adds.
NormOut fails under noise while NormAll is less affected.
A possible explanation is that a noisy target produces a large output gradient, but that gradient attenuates through backpropagation and its contribution to the full-network sum is proportionally smaller.

\paragraph{The Value of Momentum Weighting.}
The comparison of NormAll to GMC serves as an ablation study on the importance of momentum weighting for detecting learning saturation. 
In the MNIST curriculum condition (Appendix~\ref{sec:app_results}), once the three easier groups saturate, NormAll increases sampling of the hardest group only to match the other groups, resulting in slow convergence. 
GMC, in contrast, dramatically boosts sampling of the unsaturated group, producing a clear curriculum signal. 
This seems to illustrate the benefit of momentum weighting, GMC leverages it to better detect learning saturation, while gradient magnitude alone does not.

\paragraph{Limitations and Scope.}
While momentum filters oscillatory dynamics, environments with different oscillation characteristics may require different momentum decay rates.
Appendix~\ref{sec:app_beta} reports a sensitivity analysis over $\beta \in \{0.99, 0.999, 0.9999\}$ on both CIFAR-10 conditions.
GMC's qualitative behavior is stable across this range, and the main effect of a longer averaging window is to move allocations closer to uniform sampling, most visibly among groups that differ only in noise level.
This supports the reading that GMC assumes no particular noise distribution and only requires that unlearnable components lack a persistent gradient direction over the averaging window.
As with learning rate or batch size, optimal settings likely depend on the problem, and the values used here were selected through preliminary experiments.
In RL, noise can be time-dependent because a changing policy shifts the visited distribution, which suggests choosing the decay rates with the policy's rate of change in mind, favoring slower momentum where a wider range of fluctuations must cancel.
Like any intrinsic reward added to the objective, GMC can bias the policy away from the extrinsic optimum, which also holds for prediction-error curiosity and can be mitigated by annealing the intrinsic-reward weight.
As noted in Section~\ref{sec:intro}, GMC filters incoherent aleatoric variation but does not separate learning that helps the task from learning that does not, such as memorization, which we regard as a property of the target rather than a defect of the proxy and a limit shared by reward-free prediction-error curiosity.
In the controlled conditions GMC's prioritization coincides with the ordering a loss-difference signal produces, so these experiments do not exercise the decoupling between loss and learning discussed in Section~\ref{sec:intro}, and evaluating a scenario where that alignment breaks down is left for future work.
The LPM signal could be expected to help on MultiRoom as well, but it introduces a second network with its own training dynamics, so substituting it changes more than the scored quantity and makes for a less direct comparison than the ones reported here.
Integration into other architectures such as RND, and evaluation on domains with continuous control or rich visual observations, are natural extensions.
GMC scores \emph{whether learning is occurring}, methods like ICM determine \emph{what features} to measure it on, and the MultiRoom experiments show that these roles compose.

\subsection*{AI use statement}

Generative AI tools were used for language editing and \LaTeX{} formatting.
The authors reviewed and verified all AI-assisted changes and take responsibility for the final content of this work.

\subsection*{Reproducibility statement}

The method and experimental setup are described in Section~\ref{sec:method} and Section~\ref{sec:experiments}.
Appendix~\ref{sec:app_implementation} provides the architectures, optimizer settings, hyperparameters, compute resources, and other implementation details needed to reproduce the experiments.
%Code for reproducing all experiments is available at \url{https://github.com/[to-be-filled-upon-publication]}.
Code for reproducing all experiments is available at \url{https://github.com/benketriel/measuring-learning-rl}. % TODO uncomment on camera ready!

% TODO Add acknowledgments here for the camera-ready version.
% The work is partly supported by Knut and Alice Wallenberg Foundation under the Wallenberg Scholars program (Grant No. KAW2023.0327).
% Computational resources were provided by the National Academic Infrastructure for Supercomputing in Sweden (NAISS), funded by the Swedish Research Council.

%%%%%%%%%%%%%%%%%%%%%%%%%%%%%%%%%%%%%%%%%%%%%%%%%%%%%%%%%%%%%%%%
%% Bibliography
%%%%%%%%%%%%%%%%%%%%%%%%%%%%%%%%%%%%%%%%%%%%%%%%%%%%%%%%%%%%%%%%

\bibliography{iclr2027_conference}
\bibliographystyle{iclr2027_conference}

%%%%%%%%%%%%%%%%%%%%%%%%%%%%%%%%%%%%%%%%%%%%%%%%%%%%%%%%%%%%%%%%
%% Appendix
%%%%%%%%%%%%%%%%%%%%%%%%%%%%%%%%%%%%%%%%%%%%%%%%%%%%%%%%%%%%%%%%

\clearpage
\appendix

\section{Supplementary Experimental Results}\label{sec:app_results}

This section contains detailed experimental results and architectural details supporting the main paper findings.

\subsection{Derivation of the Relative Impact on Momentum Magnitude}\label{sec:app_taylor}

Equation~\ref{eq:first_order} follows from a first-order Taylor approximation of the change in momentum magnitude induced by a single sample's gradient.
Let $\mathbf{m} \in \mathbb{R}^d$ be the current momentum vector and $\mathbf{g} = \nabla_\theta \mathcal{L}_{dyn}(x) \in \mathbb{R}^d$ the gradient for sample $x$.

\paragraph{Momentum increment.}
Following Adam's definition of momentum with decay rate $\beta \in (0,1)$, the updated momentum is $\beta\mathbf{m} + (1-\beta)\mathbf{g}$, so the increment induced by $\mathbf{g}$ is:
\begin{equation}
\Delta_\mathbf{m} = \big(\beta\mathbf{m} + (1-\beta)\mathbf{g}\big) - \mathbf{m} = (\beta-1)\mathbf{m} + (1-\beta)\mathbf{g}
\label{eq:delta_m}
\end{equation}

\paragraph{First-order expansion.}
Define $f(\mathbf{x}) = \|\mathbf{m} + \mathbf{x}\|$ for $\mathbf{x} \in \mathbb{R}^d$, so that $f(\mathbf{0}) = \|\mathbf{m}\|$.
The quantity of interest is the change in momentum magnitude, which is exactly:
\begin{equation}
\Delta = f(\Delta_\mathbf{m}) - f(\mathbf{0}) = \|\mathbf{m} + \Delta_\mathbf{m}\| - \|\mathbf{m}\|
\end{equation}
Expanding $f$ around $\mathbf{x} = \mathbf{0}$ to first order:
\begin{equation}
f(\mathbf{x}) \approx f(\mathbf{0}) + \nabla f\big|_{\mathbf{x}=\mathbf{0}}^{\!\top} \mathbf{x},
\qquad
\nabla f(\mathbf{x}) = \frac{\mathbf{m} + \mathbf{x}}{\|\mathbf{m} + \mathbf{x}\|},
\qquad
\nabla f\big|_{\mathbf{x}=\mathbf{0}} = \frac{\mathbf{m}}{\|\mathbf{m}\|}
\end{equation}
Evaluating the expansion at $\mathbf{x} = \Delta_\mathbf{m}$:
\begin{equation}
\Delta \approx \frac{\mathbf{m}^{\top} \Delta_\mathbf{m}}{\|\mathbf{m}\|}
\end{equation}

\paragraph{Substituting the increment.}
Inserting Equation~\ref{eq:delta_m} and using $\mathbf{m}^{\top}\mathbf{m} = \|\mathbf{m}\|^2$:
\begin{equation}
\Delta \approx \frac{(\beta-1)\,\mathbf{m}^{\top}\mathbf{m} + (1-\beta)\,\mathbf{m}^{\top}\mathbf{g}}{\|\mathbf{m}\|}
= (\beta-1)\|\mathbf{m}\| + (1-\beta)\,\frac{\mathbf{m}^{\top} \mathbf{g}}{\|\mathbf{m}\|}
\label{eq:delta_full}
\end{equation}

\paragraph{Relative impact.}
The momentum $\mathbf{m}$ is held fixed while the current batch is scored, so the first term of Equation~\ref{eq:delta_full} is a decay term identical for every sample, and the remaining factor $(1-\beta) > 0$ is likewise common to all samples.
Neither affects how samples compare to one another: discarding them is a strictly increasing affine map of $\Delta$, which leaves the ordering of samples unchanged.
What remains is the relative impact of a gradient on the momentum magnitude:
\begin{equation}
\rho(x) = \frac{\mathbf{m}^{\top} \mathbf{g}}{\|\mathbf{m}\|}
\end{equation}
This is the scalar projection of $\mathbf{g}$ onto $\mathbf{m}$, measuring the component of the sample gradient in the direction of ongoing learning.
The absolute value in Equation~\ref{eq:r_gm} is applied to $\rho$ rather than to $\Delta$, since it is the sample-dependent part whose sign distinguishes reinforcing from opposing gradients.
The expansion is valid when $\|\Delta_\mathbf{m}\| = (1-\beta)\|\mathbf{g} - \mathbf{m}\| \ll \|\mathbf{m}\|$, which the factor $(1-\beta)$ enforces at the decay rates we use, except where an individual gradient is orders of magnitude larger than the accumulated momentum.

% \clearpage

\subsection{Robustness to Alternative Class-to-Group Assignments}\label{sec:app_repartition}

The Curriculum condition in Section~\ref{sec:controlled} uses a single fixed assignment of the ten CIFAR-10 classes to the four groups.
The prioritization order it produces could therefore reflect the visual properties of the particular classes placed in each group rather than a property of the compared signals.
To test this, we repeated the CIFAR-10 Curriculum experiment on two additional partitions whose class-to-group assignments were drawn at random (Table~\ref{table:partitions}).
The group sizes $(1,2,3,4)$ are held fixed, so the number of admissible labels per group is unchanged and only the identity of the classes varies.
Each alternative partition was run for 20 random seeds with all remaining hyperparameters as in Table~\ref{table:hyperparams}.

\begin{table}[ht]
    \centering
    \small
    \begin{tabular}{lllll}
        \toprule
        \textbf{Partition} & $G_0$ & $G_1$ & $G_2$ & $G_3$ \\
        \midrule
        Original       & $\{0\}$ & $\{1,2\}$ & $\{3,4,5\}$ & $\{6,7,8,9\}$ \\
        Randomized 1   & $\{7\}$ & $\{2,5\}$ & $\{1,6,9\}$ & $\{0,3,4,8\}$ \\
        Randomized 2   & $\{4\}$ & $\{0,8\}$ & $\{2,3,7\}$ & $\{1,5,6,9\}$ \\
        \bottomrule
    \end{tabular}
    \caption{CIFAR-10 class-to-group assignments. The two randomized partitions preserve the group sizes $(1,2,3,4)$ of the original partition and reassign the class identities at random.}
    \label{table:partitions}
\end{table}

The Uniform panel of each figure shows how the class assignment changes the empirical difficulty of the groups, and the singleton $G_0$ is the easiest in all three partitions.
Beyond that the groups reorder, and both randomized partitions are more strongly differentiated than the original one, where the three larger groups end up close together.

\begin{figure}[ht]
    \begin{center}
        \includegraphics[width=0.95\linewidth]{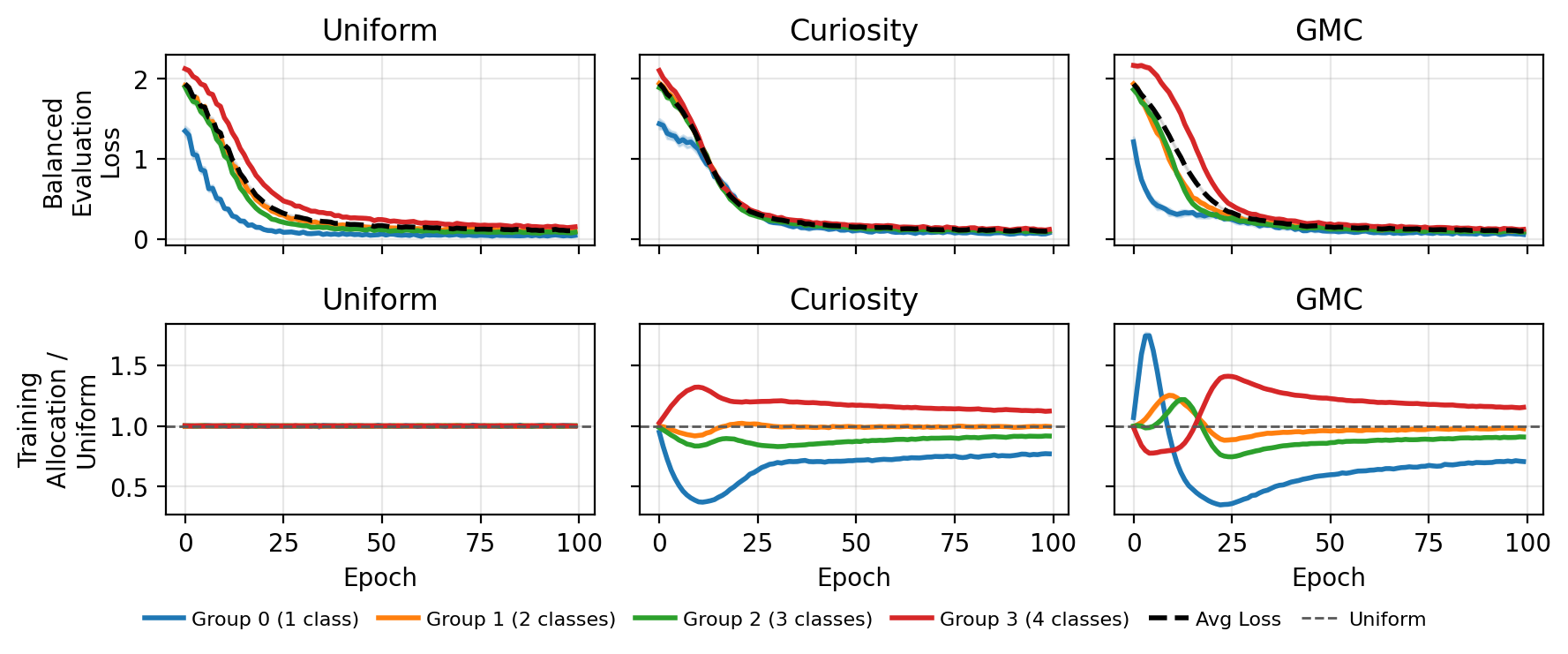}
    \end{center}
    \caption{CIFAR-10 Curriculum Condition under randomized partition 1. \textbf{Top:} Per-group Balanced Evaluation Loss. \textbf{Bottom:} Mean per-class training allocation within each group. $G_3$ is the clearly hardest group in this partition, and GMC is the only method that visits it last.}
    \label{fig:alt_shuffle_1}
\end{figure}

\paragraph{Randomized partition 1.}
Here $G_3$ is visibly harder than the other groups under Uniform sampling: its loss is the highest from the start and stays clearly above the rest for the whole run, while $G_1$ and $G_2$ track each other closely (Figure~\ref{fig:alt_shuffle_1}).
GMC visits $G_0$ first and $G_3$ last, with $G_1$ and $G_2$ in between and too close to separate reliably.
Curiosity does the opposite, allocating most to $G_3$ from the first epochs and least to $G_0$.

\begin{figure}[ht]
    \begin{center}
        \includegraphics[width=0.95\linewidth]{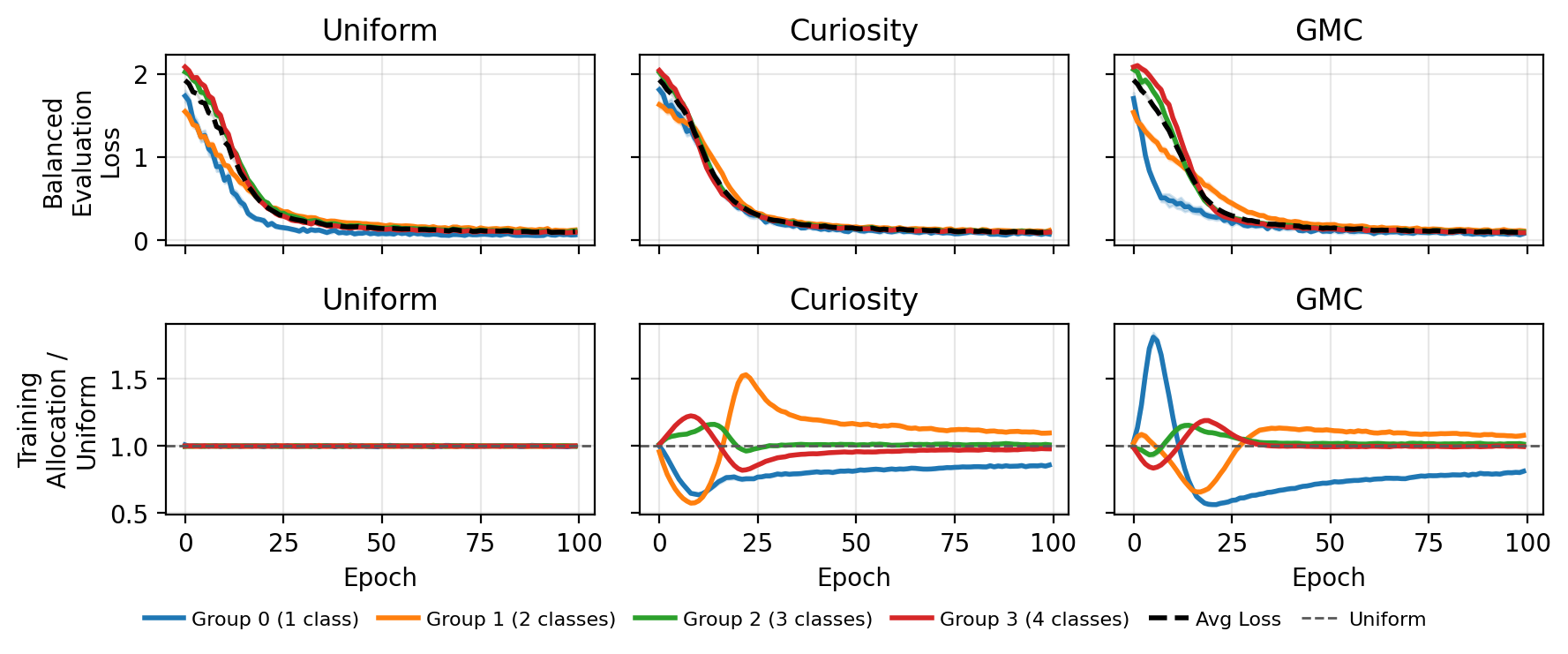}
    \end{center}
    \caption{CIFAR-10 Curriculum Condition under randomized partition 2. \textbf{Top:} Per-group Balanced Evaluation Loss. \textbf{Bottom:} Mean per-class training allocation within each group. Under Uniform sampling $G_1$ starts with the lowest loss of the multi-class groups and finishes with the highest. Curiosity turns to it only once its loss is largest, whereas GMC visits it briefly at the start and then commits to it from the middle of training onward.}
    \label{fig:alt_shuffle_2}
\end{figure}

\paragraph{Randomized partition 2.}
This partition produces a more interesting case (Figure~\ref{fig:alt_shuffle_2}).
Under Uniform sampling $G_1$ has the lowest loss of the multi-class groups at the start, yet it is overtaken by all of them later and finishes highest, so it is easy at first and hard to master.
Curiosity, following absolute loss, allocates least to $G_1$ early and only turns to it once its loss has become the largest, peaking around epoch 20 and staying elevated thereafter.
GMC's allocation for $G_1$ has two distinct phases: a small bump in the first epochs while its loss is still falling quickly, then a decline, and finally a sustained commitment from roughly epoch 35 onward as the other groups saturate and $G_1$ becomes the only group where learning is still under way.
This is the behavior the signal is intended to produce, and it is not available to a rule based on loss magnitude alone, which would have committed to $G_1$ either at the start or at the end but not at both.

\paragraph{Interpretation.}
Across all three partitions, GMC's allocation order changes together with the empirical learning dynamics of the groups rather than with their class identities, so the prioritization pattern reported in Section~\ref{sec:controlled} is not an artifact of the original assignment.
Curiosity, by contrast, remains primarily responsive to absolute prediction loss, so its prioritization begins with the groups that have the highest losses at the time, regardless of whether those groups are the ones currently improving fastest.
These are observations about which groups each signal visits and when, read off the allocation curves, and we do not claim a general correspondence between these orderings and any single scalar measure of difficulty.

% \clearpage

\subsection{MultiRoom Action Composition and Intrinsic Reward Allocation}\label{sec:app_multiroom}

\begin{figure}[ht]
    \begin{center}
        \includegraphics[width=0.98\linewidth]{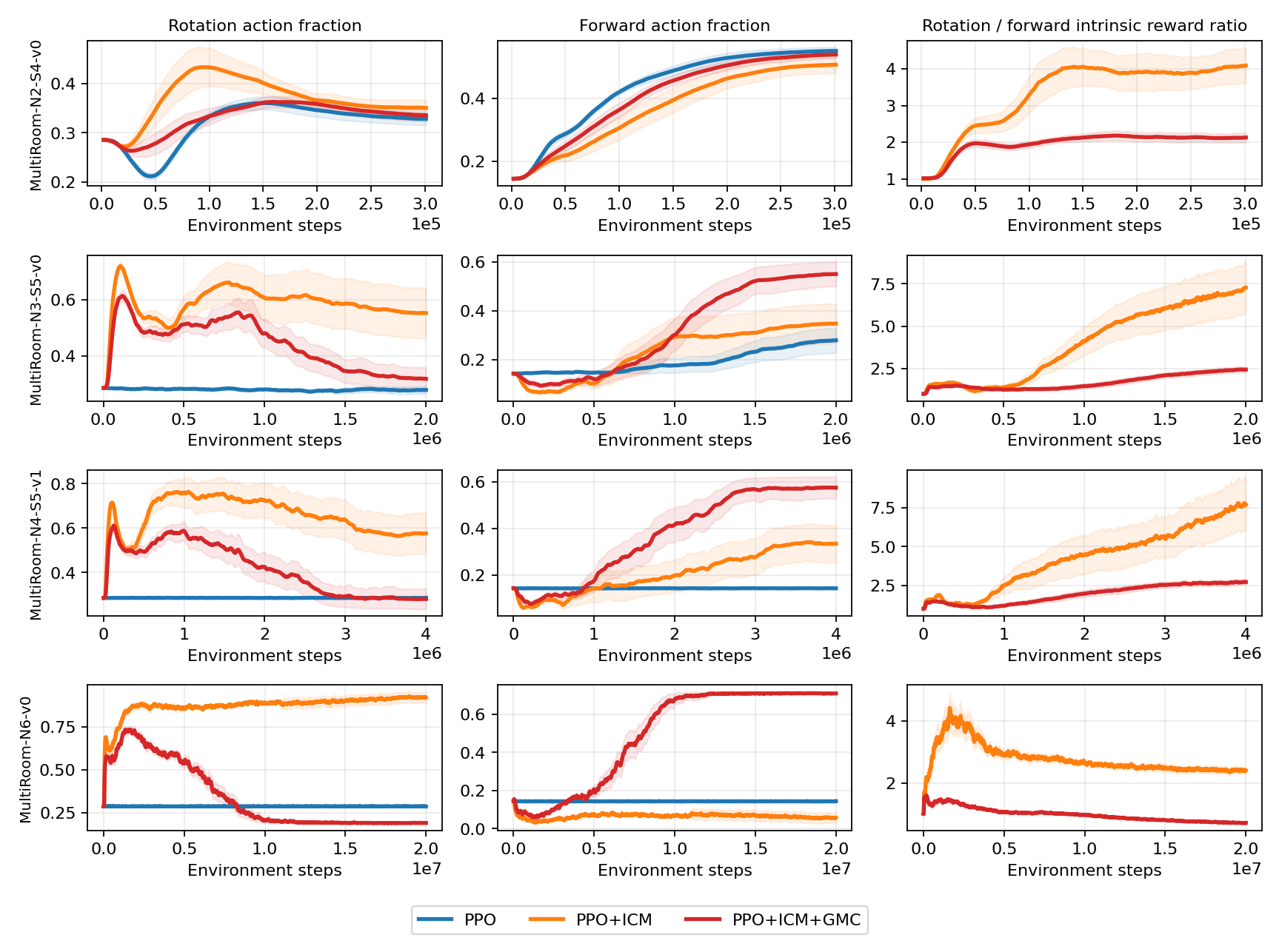}
    \end{center}
    \caption{MultiRoom behavioral diagnostics. \textbf{Columns:} fraction of actions that are rotations, fraction that are forward moves, and share of intrinsic reward received by rotation actions relative to forward moves. \textbf{Rows:} the four environments in order of increasing size. The reward-share column omits PPO, which receives no intrinsic reward.}
    \label{fig:multiroom_actions}
\end{figure}

Section~\ref{sec:minigrid} attributes ICM's declining advantage in the larger MultiRoom environments to the prediction error that rotation actions generate under partial observability.
Figure~\ref{fig:multiroom_actions} reports the behavioral quantities that this account predicts, measured over training for all three methods and all four environments.
The columns are the fraction of actions that are rotations, the fraction that are forward moves, and the share of intrinsic reward received by rotation actions relative to forward moves.
The rows are the four environments, in order of increasing size.
The reward-share column is undefined for PPO, which has no intrinsic reward.

ICM assigns a persistently larger share of its intrinsic reward to rotation than ICM+GMC does, and its action distribution remains more heavily weighted toward turning and correspondingly less toward forward movement.
The difference is present in all four environments and does not diminish with training, which is what distinguishes an irreducible residual from a transient one.
It is also largest in the six-room environment, where ICM turns for the great majority of its steps while ICM+GMC turns less than in any smaller environment, matching the return curves in Figure~\ref{fig:multiroom_return}.

% \clearpage

\subsection{Ablation: GMC Variants}\label{sec:app_ablation}

We compare three variants of the GMC signal:

\begin{enumerate}
    \item \textbf{GMC}: $d^{-1/2}\sum_i \lvert g_i \hat m_i/s_i \rvert$
    \item \textbf{Dot Product}: $d^{-1/2}\sum_{T}\left\lvert\langle g_T,\hat m_T/s_T\rangle\right\rvert$
    \item \textbf{Cosine Similarity}: $d^{-1/2}\sum_T \left\lvert\cos(g_T,\hat m_T/s_T)\right\rvert$
\end{enumerate}
Here $s_i=\max(\hat v_i,\alpha^2\hat q_i)$ as in Equation~\ref{eq:gmc_scale}, and $T$ ranges separately over every weight and bias tensor.

\paragraph{Theoretical Differences.}
The Dot Product variant aggregates products within each parameter tensor before taking magnitude, permitting cancellation within a tensor while keeping weight and bias contributions from each layer separate.
In contrast, GMC applies the absolute value elementwise, allowing parameters that change in different directions to contribute independently. 
As a result, GMC captures simultaneous multi-dimensional changes rather than collapsing them into a single axis.
The Cosine Similarity variant computes an absolute cosine separately for every weight and bias tensor, using the same normalized momentum $\hat m/s$ as the other variants. It is scale-invariant within each tensor and measures only directional alignment, discarding magnitude information.

\begin{figure}[ht]
    \begin{center}
        \includegraphics[width=0.95\linewidth]{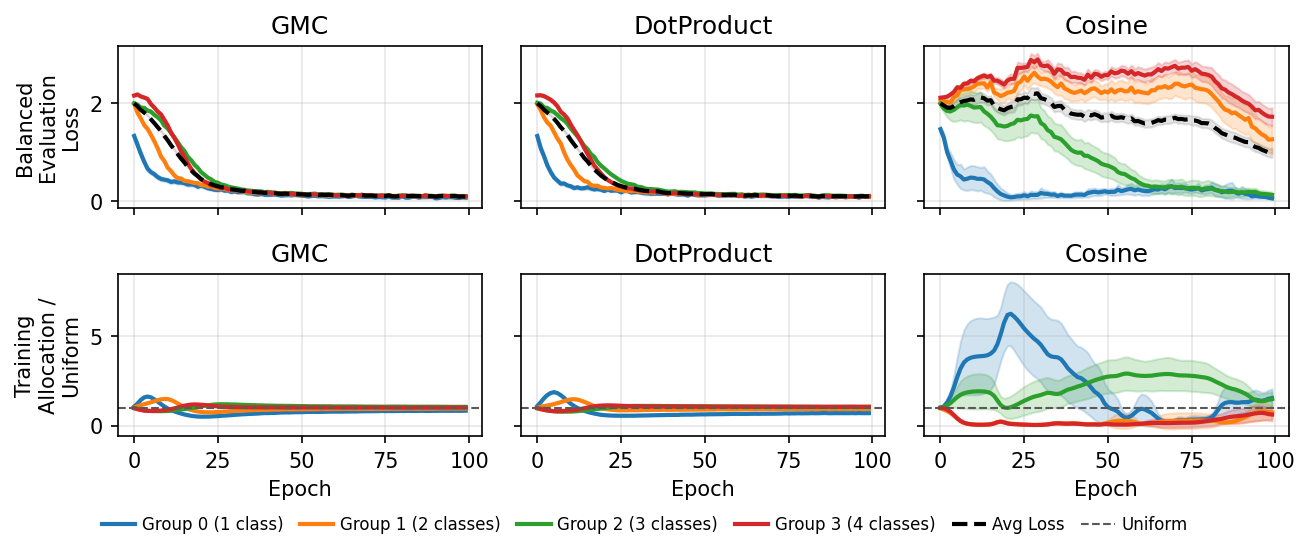}
    \end{center}
    \caption{CIFAR-10 Curriculum Condition for the three GMC variants, one per column: GMC, Dot Product, and Cosine Similarity. \textbf{Top row:} Per-group Balanced Evaluation Loss. \textbf{Bottom row:} Mean per-class training allocation within each group. GMC and Dot Product produce broadly similar allocations, while Cosine fluctuates throughout.}
    \label{fig:abs_ablation}
\end{figure}

\paragraph{Empirical Observations.}
Cosine Similarity is clearly unsuitable: it tends to oversample the easiest group early in training and exhibits fluctuating sampling weights throughout (Figure~\ref{fig:abs_ablation}).
By construction, cosine normalization amplifies even residual gradient alignment once magnitudes shrink near convergence, injecting noise into the signal precisely when it should vanish.

Dot Product and GMC yield broadly similar sampling behavior.
Given these results, both GMC and Dot Product are viable.
We default to GMC because its elementwise absolute value preserves independent per-parameter contributions across multiple simultaneous directions of change, whereas Dot Product permits cancellation among parameters within each tensor.
This ablation does not establish that either viable variant is superior in more complex domains.

% \clearpage

\subsection{Sensitivity to the EMA Decay Rate}\label{sec:app_beta}

GMC maintains three exponentially-weighted moving averages: the momentum $m_i$, second moment $v_i$, and squared-parameter scale $q_i$. In this sweep their shared decay rate $\beta$ sets the length of the averaging window.
To quantify how strongly the signal depends on this choice, we repeated the CIFAR-10 Curriculum and Noise conditions with $\beta_m = \beta_v \in \{0.99, 0.999, 0.9999\}$, using 10 random seeds per configuration.
All remaining hyperparameters are as in Table~\ref{table:hyperparams}.

\begin{figure}[ht]
    \begin{center}
        \includegraphics[width=0.95\linewidth]{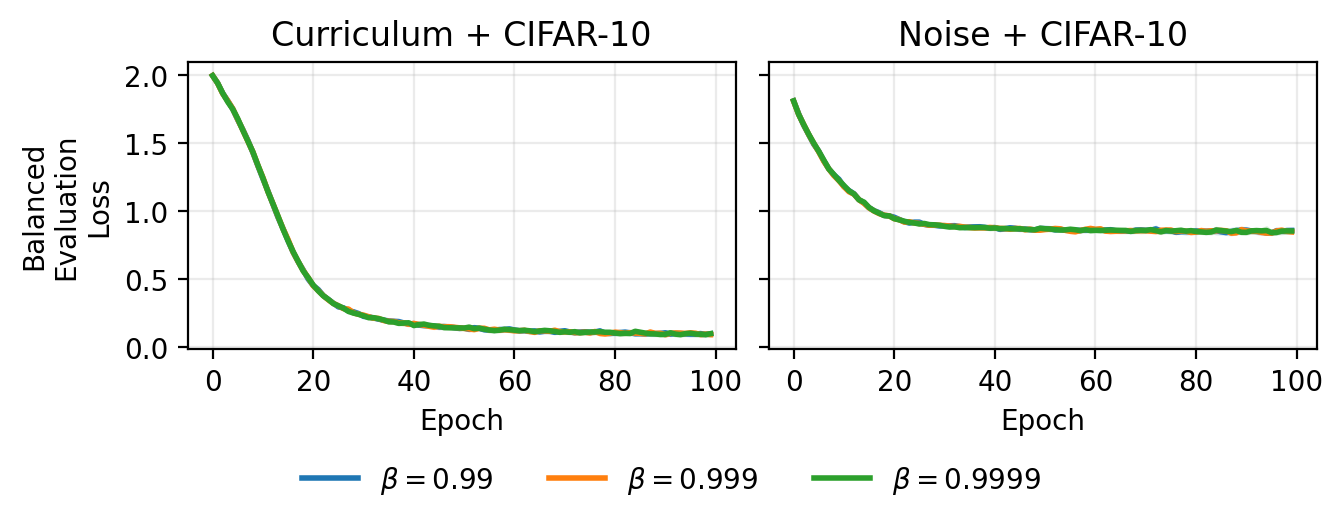}
    \end{center}
    \caption{Balanced Evaluation Loss averaged over all classes for the three decay rates. The three curves are indistinguishable in both conditions.}
    \label{fig:beta_loss_cifar}
\end{figure}

\begin{figure}[ht]
    \begin{center}
        \includegraphics[width=0.95\linewidth]{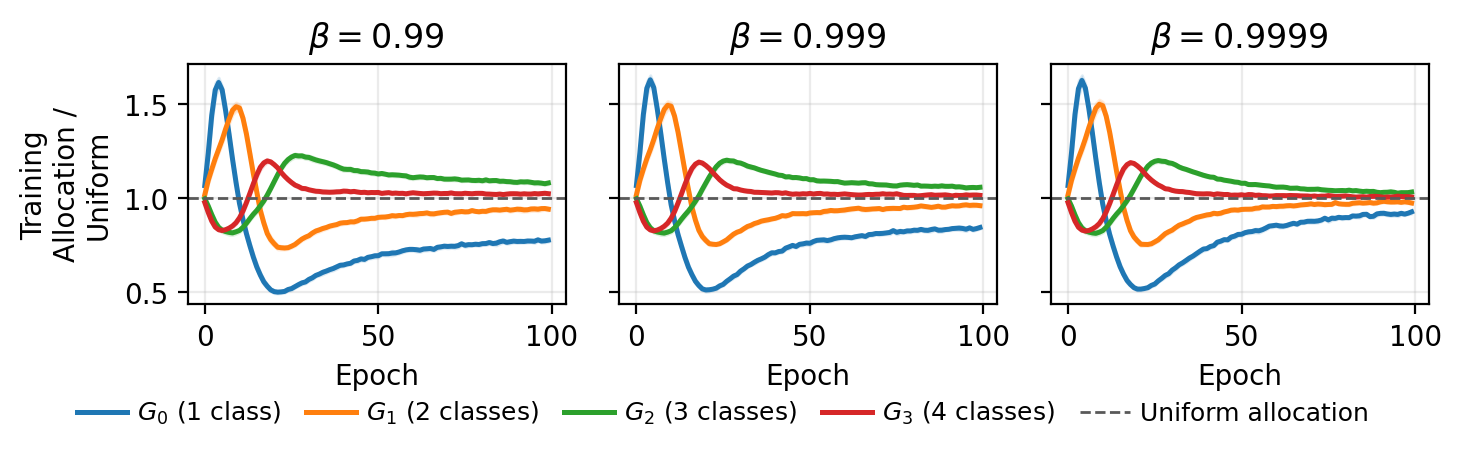}
    \end{center}
    \caption{GMC decay-rate sensitivity on the CIFAR-10 Curriculum Condition, mean per-class training allocation within each group. The visitation peaks occur in the same order for all three decay rates, and the allocations contract toward uniform sampling as $\beta$ increases.}
    \label{fig:beta_curriculum_cifar}
\end{figure}

\begin{figure}[ht]
    \begin{center}
        \includegraphics[width=0.95\linewidth]{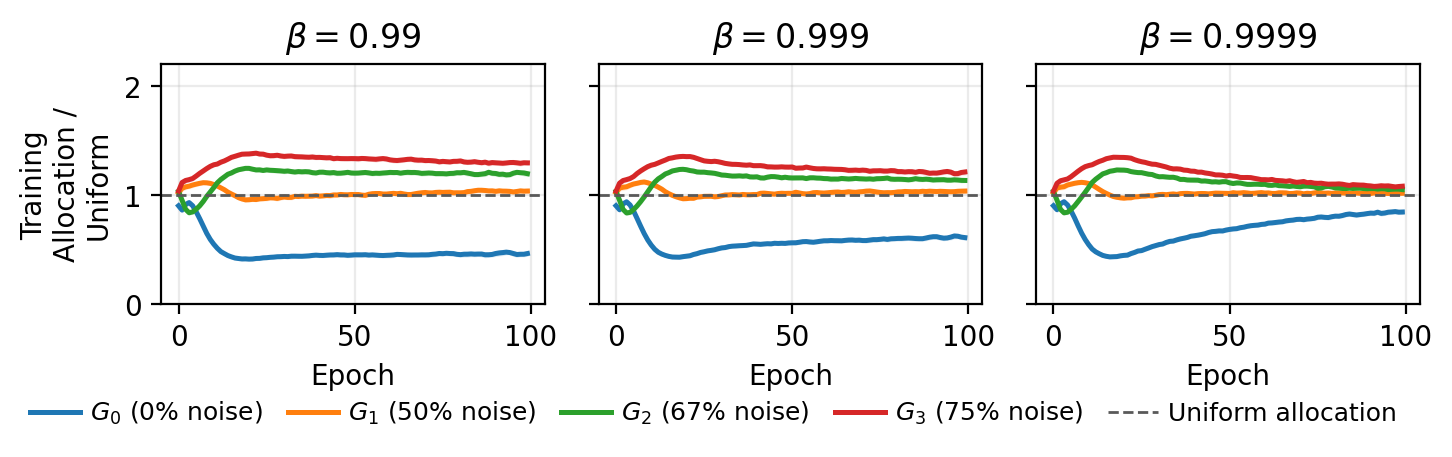}
    \end{center}
    \caption{GMC decay-rate sensitivity on the CIFAR-10 Noise Condition, mean per-class training allocation within each group. The three noisy groups draw closer together as $\beta$ increases, while the gap between them and the noise-free $G_0$ remains visible at every decay rate.}
    \label{fig:beta_noise_cifar}
\end{figure}

\paragraph{Convergence.}
The overall loss is unaffected by the decay rate (Figure~\ref{fig:beta_loss_cifar}).
In both conditions the three curves are indistinguishable over the whole of training, so the choice of $\beta$ within this range has no measurable effect on convergence speed or final loss.
What it does affect is the allocation.

\paragraph{Curriculum Condition.}
The ordering of the visitation peaks is the same for all three values of $\beta$, namely $G_0 \rightarrow G_1 \rightarrow G_3 \rightarrow G_2$ (Figure~\ref{fig:beta_curriculum_cifar}), so the prioritization reported in Section~\ref{sec:controlled} is not an artifact of the particular decay rate used there.
The peak heights are also largely unchanged.
The visible effect of a longer window is a contraction of the later allocations toward uniform sampling, resembling the effect of raising the entropy-regularization weight $\lambda_{\text{ent}}$.

\paragraph{Noise Condition.}
The same contraction appears here where the four groups differ only in noise level (Figure~\ref{fig:beta_noise_cifar}).
As $\beta$ increases, the separation between the three noisy groups shrinks, and by $\beta=0.9999$ their final allocations nearly coincide.
The gap between those groups and the noise-free $G_0$ persists at every decay rate, although it too narrows.
This is the behavior expected from the mechanism described in Section~\ref{sec:intro}.
A larger $\beta$ corresponds to a longer effective averaging window, over which inconsistent gradient components have more opportunity to cancel, which makes GMC less sensitive to transient differences between groups that differ only in noise level.
It is the empirical counterpart of the assumption stated in the Limitations, that GMC requires only that unlearnable components lack a persistent gradient direction over the averaging window.

\paragraph{Summary.}
Sensitivity over the tested range is limited: convergence is unaffected and the prioritization order is unchanged, while a longer window moves allocations toward uniform sampling and slightly improves noise robustness.
We use $\beta=0.999$ in both the controlled experiments and the MultiRoom experiments.

% \clearpage

\subsection{Implementation Details}\label{sec:app_implementation}

\begin{table}[ht]
    \centering
    \small
    \begin{tabular}{|l|c|l|}
        \hline
        \textbf{Component} & \textbf{Type} & \textbf{Configuration} \\
        \hline
        \textbf{Dynamics Model} & MLP & Input: $32 \times 32 \times 3$ (CIFAR-10) / $28 \times 28 \times 1$ (MNIST) \\
        (Classifier) & & Flattened to 3072 / 784 neurons \\
        & & Hidden Layer 1: 512 neurons, ReLU \\
        & & Hidden Layer 2: 512 neurons, ReLU \\
        & & Output Layer: 10 neurons (logits) \\
        & & Loss: Cross-Entropy \\
        \hline
        \textbf{Policy Network} & MLP & Input: 4 (groups) or 10 (classes) neurons of independent uniform random noise \\
        (Actor) & & Hidden Layer 1: 512 neurons, ReLU \\
        & & Hidden Layer 2: 512 neurons, ReLU \\
        & & Output Layer: 4 (groups) or 10 (classes) neurons (logits) \\
        & & Action: $\text{softmax}(\text{output})$ for group selection \\
        \hline
    \end{tabular}
    \caption{Network architectures for dynamics model and policy. Both use identical hidden dimension (512) and ReLU activations. The dynamics model predicts class labels, the policy provides sampling probabilities at each step.}
    \label{table:architecture}
\end{table}

\begin{table}[ht]
    \centering
    \small
    \begin{tabular}{|l|l|}
        \hline
        \textbf{Hyperparameter} & \textbf{Value} \\
        \hline
        \multicolumn{2}{|c|}{\textbf{Optimization}} \\
        \hline
        Dynamics Model & Adam, learning rate $1 \times 10^{-3}$ \\
        Policy Network & Adam, learning rate $5 \times 10^{-6}$, $\beta_1 = 0.99$ \\
        \hline
        \multicolumn{2}{|c|}{\textbf{Training}} \\
        \hline
        Batch Size & 256 \\
        Evaluation Batches per Epoch & 100 \\
        Num Epochs (MNIST Curriculum) & 400 \\
        Num Epochs (other conditions) & 100 \\
        Entropy Regularization Weight ($\lambda_{\text{ent}}$) & 0.1 \\
        Intrinsic Signal Weight & 1.0 \\
        Signal-scale EMA Decay & 0.9 \\
        Signal-scale Calibration Window & First 10\% of training \\
        \hline
        \multicolumn{2}{|c|}{\textbf{GMC}} \\
        \hline
        Momentum Decay ($\beta_m$) & 0.999 \\
        Second-Moment Decay ($\beta_v$) & 0.999 \\
        Squared-Parameter Decay ($\beta_q=\beta_v$) & 0.999 \\
        Parameter-Floor Scale ($\alpha$) & $1 \times 10^{-2}$ \\
        \hline
        \multicolumn{2}{|c|}{\textbf{Data}} \\
        \hline
        Random Seeds & 0-19 (20 total) \\
        \hline
    \end{tabular}
    \caption{Training hyperparameters for Signal Evaluation under Controlled Difficulty and Noise experiments (CIFAR-10, MNIST).
    The policy is updated after every batch. Settings not listed are library defaults.
    Policy network receives lower learning rate than dynamics model to ensure stable learning.
    Momentum decay parameters for GMC use longer averaging windows than Adam's default $\beta_1$, capturing more extended learning history.}
    \label{table:hyperparams}
\end{table}

\begin{table}[ht]
    \centering
    \small
    \begin{tabular}{|l|l|}
        \hline
        \textbf{Hyperparameter} & \textbf{Value} \\
        \hline
        \multicolumn{2}{|c|}{\textbf{Environments}} \\
        \hline
        Environment 1 & MiniGrid-MultiRoom-N2-S4-v0, 300{,}000 steps \\
        Environment 2 & MiniGrid-MultiRoom-N3-S5-v0 (locally registered), 2{,}000{,}000 steps \\
        Environment 3 & MiniGrid-MultiRoom-N4-S5-v1, 4{,}000{,}000 steps \\
        Environment 4 & MiniGrid-MultiRoom-N6-v0, 20{,}000{,}000 steps \\
        Observation & Partial, egocentric (as distributed) \\
        Environment Modifications & None \\
        \hline
        \multicolumn{2}{|c|}{\textbf{PPO}} \\
        \hline
        Learning Rate & $2.5 \times 10^{-4}$, linearly annealed \\
        Discount Factor ($\gamma$) & 0.99 \\
        GAE $\lambda$ & 0.95 \\
        Entropy Coefficient & 0.01 \\
        PPO Epochs per Rollout & 4 \\
        Mini-Batch Size & 512 \\
        Parallel Environments & 16 \\
        Rollout Length per Environment & 128 \\
        Hidden Dimension & 256 \\
        \hline
        \multicolumn{2}{|c|}{\textbf{ICM}} \\
        \hline
        Feature Dimension & 128 \\
        Learning Rate & $3 \times 10^{-4}$ \\
        Epochs per Rollout & 2 \\
        Mini-Batch Size & 256 \\
        Forward / Inverse Loss Weight & 0.2 / 0.8 \\
        \hline
        \multicolumn{2}{|c|}{\textbf{Intrinsic Reward (ICM and ICM+GMC)}} \\
        \hline
        Intrinsic Reward Coefficient & $5 \times 10^{-4}$ \\
        \hline
        \multicolumn{2}{|c|}{\textbf{Data}} \\
        \hline
        Random Seeds & 0-49 (50 total) \\
        \hline
    \end{tabular}
    \caption{Hyperparameters for the MiniGrid MultiRoom reinforcement learning experiments.
    The environments are used exactly as distributed, with no modified observations or rewards, and each is run for the step budget listed above.
    The distributed set contains no three-room instance, so we register one locally with the same entry point as the other variants and three rooms of maximum size 5.
    Only settings that were chosen for these experiments are listed: the GMC decay rates and parameter-floor scale are unchanged from Table~\ref{table:hyperparams}, and the remaining PPO and ICM settings are the commonly used values of \citet{schulman2017proximal} and \citet{pathak2017curiosity}.
    ICM and ICM+GMC use the same architecture, the same dynamics model hyperparameters, and the same intrinsic reward coefficient.
    PPO hyperparameters are identical across all three methods.}
    \label{table:hyperparams_minigrid}
\end{table}

\paragraph{Hyperparameter Selection.}
PPO hyperparameters follow widely-used defaults \citep{schulman2017proximal, schulman2015high}. Preliminary experiments confirmed stable training with these values across all methods.
GMC-specific parameters (the moment decay rates and parameter-floor scale) were chosen based on the theoretical motivation that longer averaging windows smooth noise more effectively and that the denominator should remain meaningful when gradients become small, and were validated through preliminary experiments to confirm stable signal behavior.
ICM hyperparameters were set to values from the original work \citep{pathak2017curiosity}.
Controlled experiment hyperparameters (Table~\ref{table:hyperparams}) were similarly determined through initial experimentation, selecting values that produced stable learning dynamics across all methods.
A sensitivity analysis of GMC's decay rates is reported in Appendix~\ref{sec:app_beta}.

\paragraph{Computational Complexity.}
For $d$ scored parameters, GMC stores a momentum, a gradient second moment, and a squared-parameter moving average per parameter and computes an $O(d)$ elementwise product and sum.
This is the same asymptotic order as the backward pass, though the current implementation has practical time and memory overhead from obtaining per-sample gradients.
The momentum $m_i$, second moment $v_i$, and squared-parameter average $q_i$ are exponentially-weighted moving averages of the batch-mean gradient $\bar g_i$ and the parameter value $\theta_i$:
\begin{equation}
m_i \leftarrow \beta_m m_i + (1-\beta_m)\bar g_i,\quad
v_i \leftarrow \beta_v v_i + (1-\beta_v)\bar g_i^2,\quad
q_i \leftarrow \beta_v q_i + (1-\beta_v)\theta_i^2.
\end{equation}
The bias-corrected values $\hat m_i$, $\hat v_i$, and $\hat q_i$ from prior batches are used to score the current batch with denominator $\max(\hat v_i,\alpha^2\hat q_i)$, after which the current batch statistics are inserted. We use $\beta_m=\beta_v=0.999$ and $\alpha=10^{-2}$ in the controlled experiments. Equal momentum and second-moment decay rates imply $\hat m_i^2\leq\hat v_i$ by Jensen's inequality, while the parameter-relative floor further prevents amplification when $\hat v_i$ is very small.

GMC maintains its own momentum, second-moment, and parameter-scale estimates for scoring.
These read-only statistics do not interact with the optimizer's update rule or apply momentum twice.
We keep them separate to allow decay rates different from Adam's defaults ($\beta_1=0.9$, $\beta_2=0.999$).
With matched rates, the optimizer's own estimates could be reused.

We normalize GMC by $1/\sqrt d$, where $d$ is the number of scored parameters. In the controlled experiments all non-uniform methods, including NormOut, NormAll, and DeltaLoss, then use the shared early-signal calibration described below, with no method-specific scaling constants.

\paragraph{Signal Scale Calibration.}
In the controlled experiments the policy-gradient term scales with the magnitude of $r_{\text{int}}$, so a signal whose raw values are large relative to $\lambda_{\text{ent}}$ would sharpen its allocation for reasons unrelated to what it measures.
The six methods produce values on very different scales, and to compare them under a single shared $\lambda_{\text{ent}}$ we track an EMA (decay 0.9) of each signal's batch-mean magnitude during the first 10\% of training, divide the signal by that running value, and then freeze the divisor for the rest of the run.
Freezing matters, since a divisor that kept updating would normalize away the decay of the signal as learning saturates, which is precisely the behavior under study.
This is a device for fair comparison rather than a component of the method.

All experiments were run using PyTorch 2.0.1 on Ubuntu with an NVIDIA GeForce RTX 3080 GPU. The MNIST and CIFAR experiments took approximately 30 minutes per seed. Curriculum+MNIST took approximately four times longer, as it was run for 400 rather than 100 analyzed epochs, requiring about 120 minutes per seed. NormAll increased runtime, taking about 408 minutes on Curriculum+MNIST and about 90 minutes on the remaining MNIST/CIFAR experiments.
For MultiRoom each different configuration ran a different number of steps. 
For all of them PPO ran roughly at 3000 steps per second, ICM at 2600, and ICM+GMC at 2500. However, with a PyTorch native reimplementation of GMC these differences would most probably be negligible.

% \clearpage

\subsection{Complete Results Across All Conditions}

The main text reports the per-group results for Uniform, Curiosity and GMC on CIFAR-10.
Here we provide the overall loss of all six methods on both datasets, and the per-group losses and allocations for the remaining combinations of dataset, condition and method.
Table~\ref{tab:normalized-loss-auc} gives the summary metric of Table~\ref{tab:auc_cifar} for both datasets and additionally for the two GMC variants of Appendix~\ref{sec:app_ablation}.

\begin{table}[ht]
    \centering
    \small
    \begin{tabular}{lrrrrr}
        \toprule
        & \multicolumn{2}{c}{Curriculum} & \multicolumn{2}{c}{Noise} & \\
        \cmidrule(lr){2-3} \cmidrule(lr){4-5}
        Method & MNIST & CIFAR-10 & MNIST & CIFAR-10 & Total \\
        \midrule
        Uniform & $1.00 \pm 0.01$ & $1.00 \pm 0.01$ & $\mathbf{1.00 \pm 0.00}$ & $1.00 \pm 0.00$ & $4.00 \pm 0.01$ \\
        Curiosity & $\mathbf{0.92 \pm 0.01}$ & $\mathbf{0.99 \pm 0.00}$ & $1.13 \pm 0.00$ & $1.23 \pm 0.00$ & $4.27 \pm 0.01$ \\
        GMC & $0.96 \pm 0.01$ & $1.01 \pm 0.00$ & $\mathbf{1.00 \pm 0.00}$ & $\mathbf{0.99 \pm 0.00}$ & $\mathbf{3.96 \pm 0.01}$ \\
        NormOut & $0.94 \pm 0.01$ & $1.00 \pm 0.00$ & $1.13 \pm 0.00$ & $1.34 \pm 0.00$ & $4.41 \pm 0.01$ \\
        NormAll & $1.02 \pm 0.01$ & $1.00 \pm 0.00$ & $1.01 \pm 0.00$ & $1.07 \pm 0.00$ & $4.10 \pm 0.01$ \\
        DeltaLoss & $1.00 \pm 0.01$ & $1.14 \pm 0.01$ & $\mathbf{1.00 \pm 0.00}$ & $1.02 \pm 0.00$ & $4.16 \pm 0.01$ \\
        DotProduct & $0.97 \pm 0.01$ & $1.02 \pm 0.00$ & $\mathbf{1.00 \pm 0.00}$ & $1.00 \pm 0.00$ & $3.99 \pm 0.01$ \\
        Cosine & $5.30 \pm 1.59$ & $4.64 \pm 0.19$ & $1.04 \pm 0.00$ & $1.78 \pm 0.02$ & $12.76 \pm 1.60$ \\
        \bottomrule
    \end{tabular}
    \caption{Normalized area under the overall Balanced Evaluation Loss curve, lower is better. Values are scaled so that Uniform is $1.00$ in each column. The best mean in each column is shown in bold. The Total column sums the four conditions and is reported only as a compact overall ordering. DotProduct and Cosine are the GMC variants of Appendix~\ref{sec:app_ablation}.}
    \label{tab:normalized-loss-auc}
\end{table}

\begin{figure}[ht]
    \centering
    \includegraphics[width=0.95\linewidth]{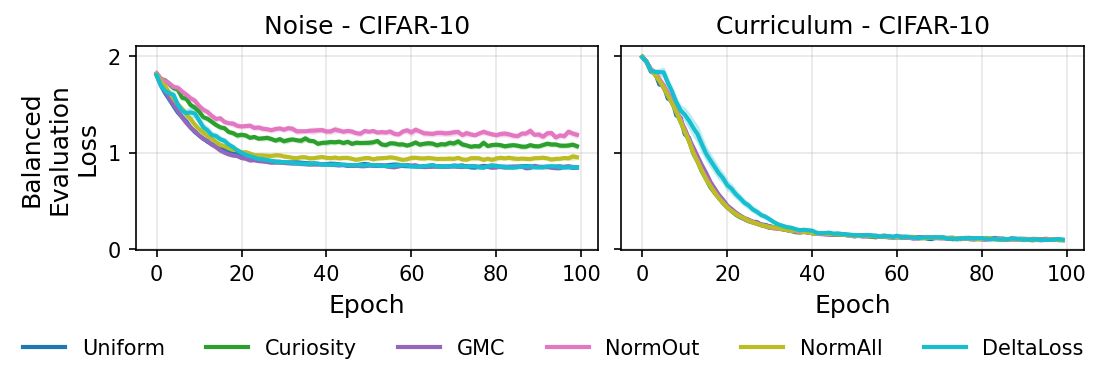}
    \caption{CIFAR-10 Balanced Evaluation Loss averaged over all classes, for all six methods in the Noise (left) and Curriculum (right) conditions.}
    \label{fig:all_conditions}
\end{figure}

\begin{figure}[ht]
    \centering
    \includegraphics[width=0.95\linewidth]{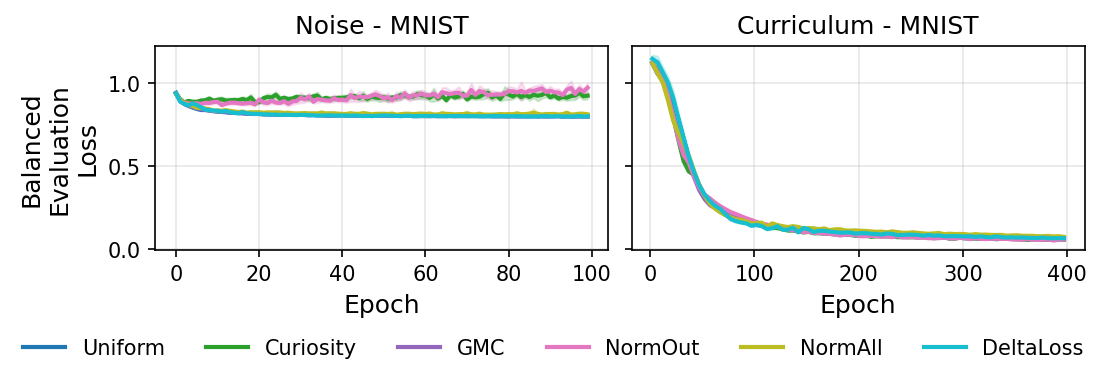}
    \caption{MNIST Balanced Evaluation Loss averaged over all classes, for all six methods in the Noise (left) and Curriculum (right) conditions. The split is the same as on CIFAR-10 (Figure~\ref{fig:all_conditions}).}
    \label{fig:all_conditions_mnist}
\end{figure}

\begin{figure}[ht]
    \begin{center}
        \includegraphics[width=0.95\linewidth]{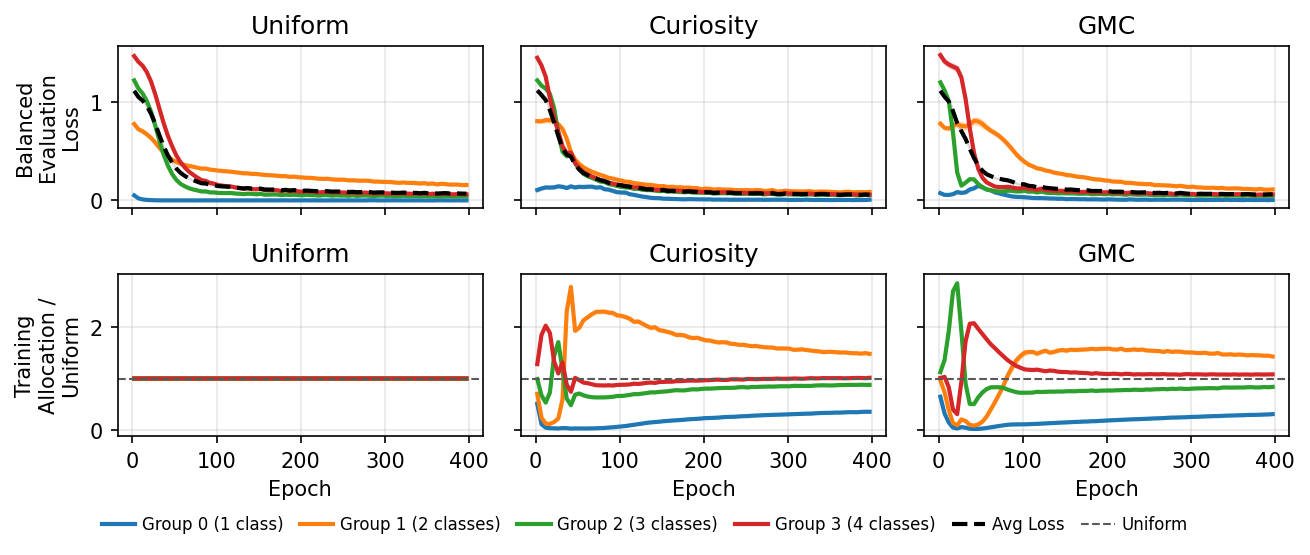}
    \end{center}
    \caption{MNIST Curriculum Condition. \textbf{Top:} Per-group Balanced Evaluation Loss. \textbf{Bottom:} Mean per-class training allocation within each group. MNIST shows larger difficulty differences between groups than CIFAR-10, producing more pronounced prioritization differences. GMC peaks in the order $G_2\!\rightarrow\!G_3\!\rightarrow\!G_1$, while Curiosity prioritizes by initial loss magnitude ($G_3\!\rightarrow\!G_2\!\rightarrow\!G_1$). $G_0$ is left out of both orderings: it is the easiest group and its allocation stays low for both methods without forming a distinct peak, so it has no well-defined position in the sequence.}
    \label{fig:curriculum_mnist}
\end{figure}

\begin{figure}[ht]
    \begin{center}
        \includegraphics[width=0.95\linewidth]{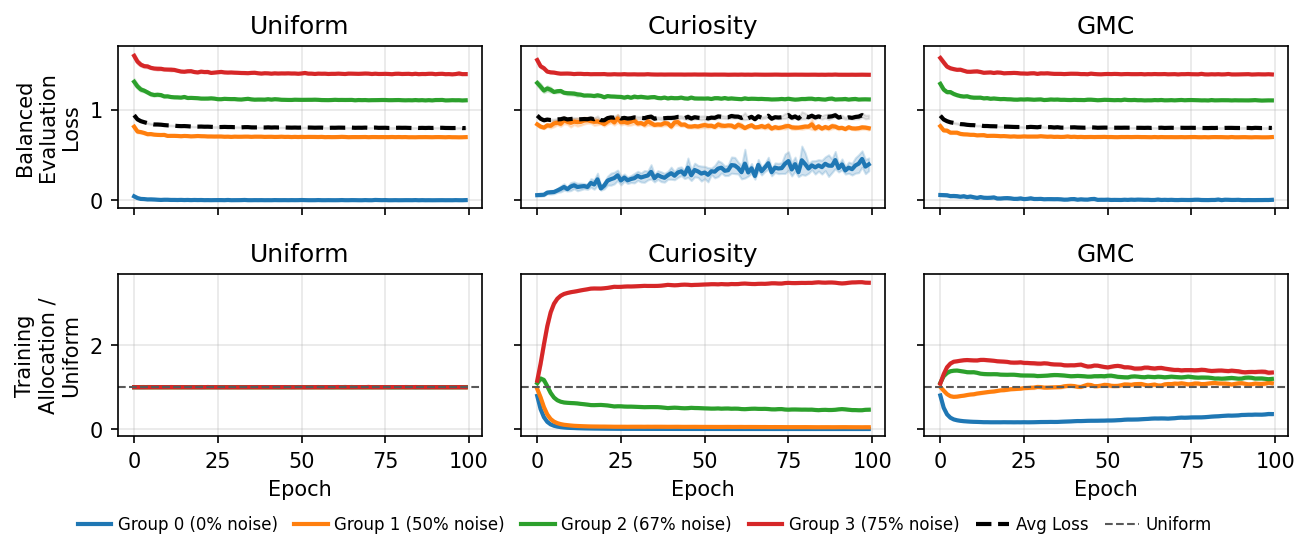}
    \end{center}
    \caption{MNIST Noise Condition. \textbf{Top:} Per-group Balanced Evaluation Loss. \textbf{Bottom:} Mean per-class training allocation within each group. Like CIFAR-10, Curiosity concentrates on the noisiest group while neglecting others. GMC maintains stable allocation.}
    \label{fig:noise_mnist}
\end{figure}

\begin{figure}[ht]
    \begin{center}
        \includegraphics[width=0.95\linewidth]{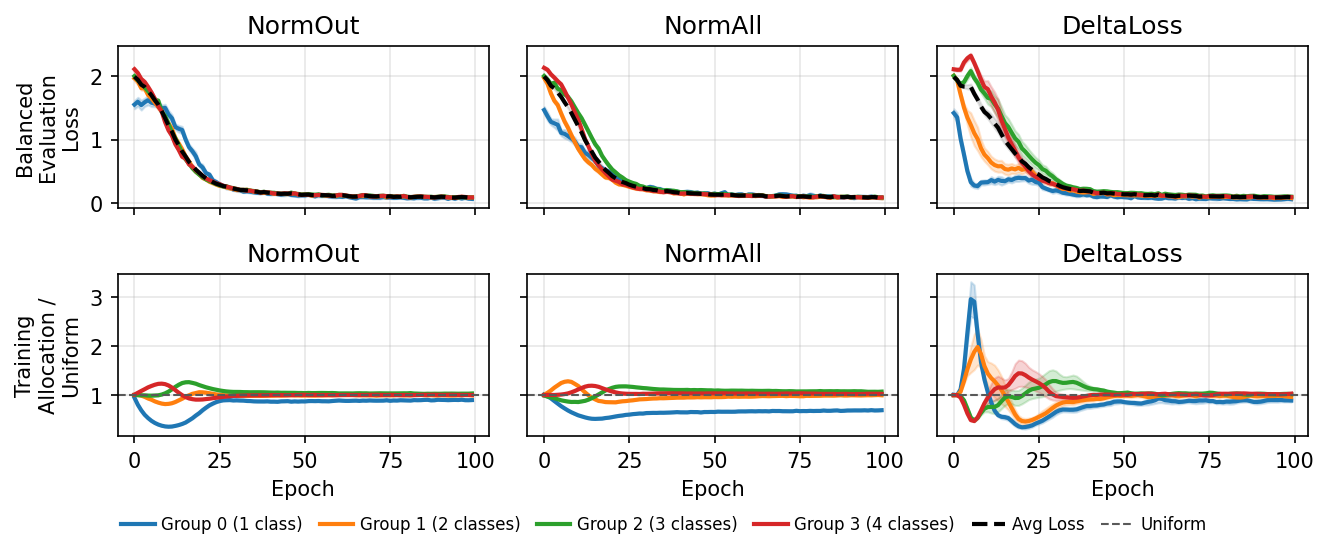}
    \end{center}
    \caption{CIFAR-10 Curriculum Condition for NormOut, NormAll, and DeltaLoss. \textbf{Top row:} Per-group Balanced Evaluation Loss. \textbf{Bottom row:} Mean per-class training allocation within each group. NormOut shows prioritization similar to Curiosity and NormAll to GMC, but both exhibit slower convergence. DeltaLoss peaks in the order $G_0 \rightarrow G_1 \rightarrow G_3 \rightarrow G_2$, the same order as GMC in Figure~\ref{fig:curriculum_cifar}, but its allocation is oscillatory and has much wider confidence intervals, reflecting the variance of consecutive-epoch loss estimates.}
    \label{fig:curriculum_cifar_variants}
\end{figure}

\begin{figure}[ht]
    \begin{center}
        \includegraphics[width=0.95\linewidth]{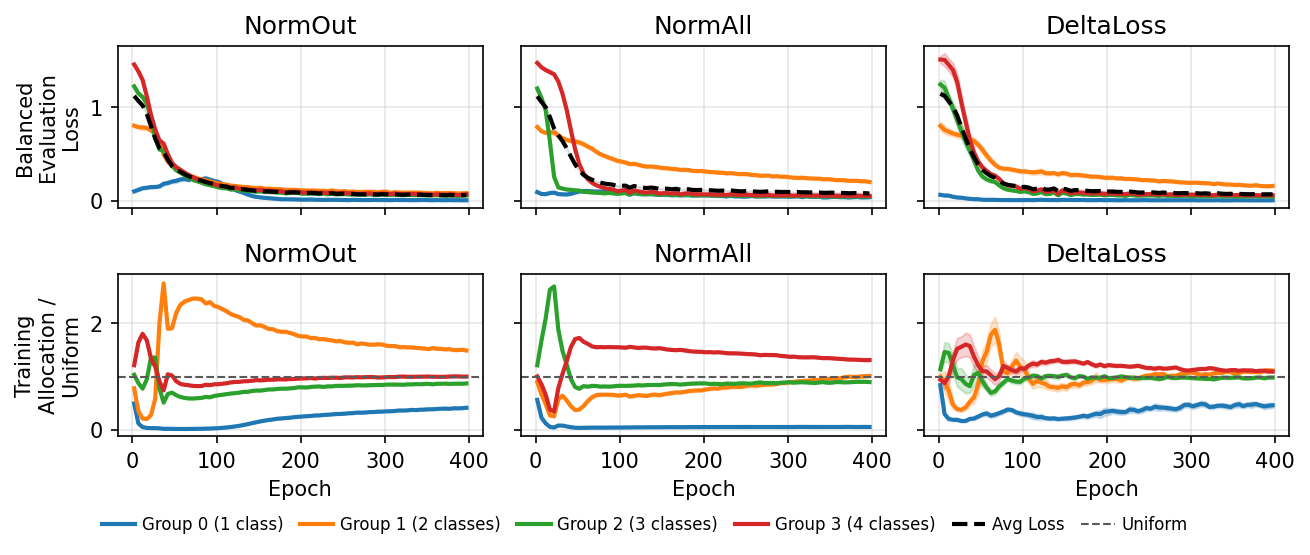}
    \end{center}
    \caption{MNIST Curriculum Condition for NormOut, NormAll, and DeltaLoss. \textbf{Top row:} Per-group Balanced Evaluation Loss. \textbf{Bottom row:} Mean per-class training allocation within each group. NormAll fails to prioritize the hardest group after easier groups saturate, leading to slower learning. These results emphasize the role of momentum weighting. DeltaLoss exhibits pronounced oscillations due to high variance in consecutive-epoch loss differences, which harm the convergence speed.}
    \label{fig:curriculum_mnist_variants}
\end{figure}

\begin{figure}[ht]
    \begin{center}
        \includegraphics[width=0.95\linewidth]{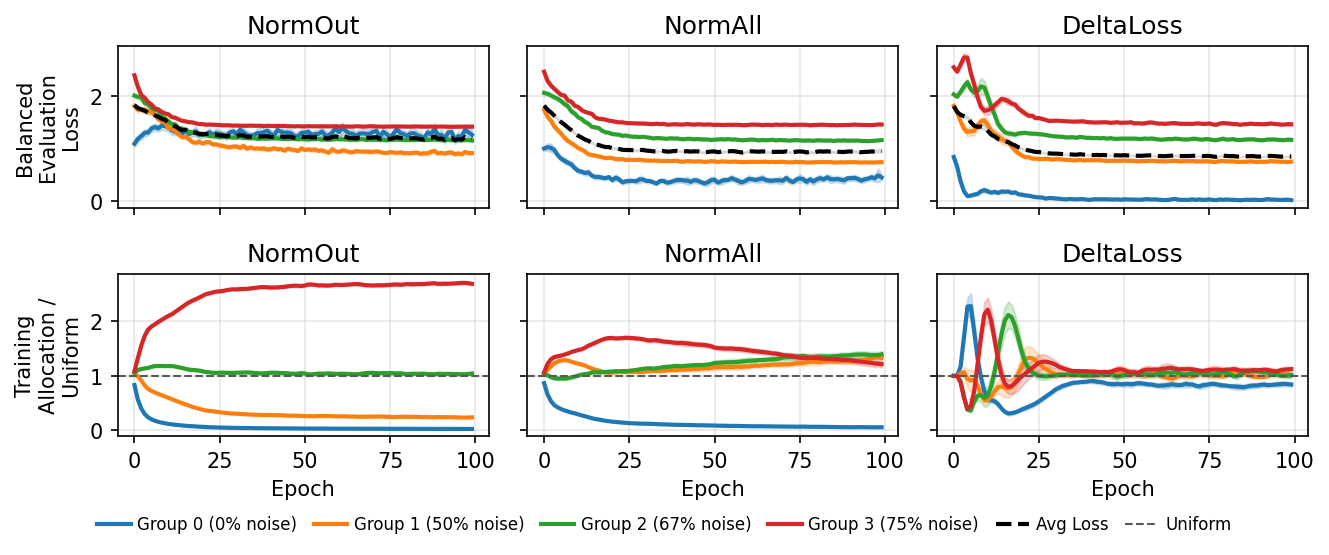}
    \end{center}
    \caption{CIFAR-10 Noise Condition for NormOut, NormAll, and DeltaLoss. \textbf{Top row:} Per-group Balanced Evaluation Loss. \textbf{Bottom row:} Mean per-class training allocation within each group. NormOut is strongly affected by noise. NormAll's allocation is also affected, but less severely, highlighting the contribution of measuring the gradient over all parameters rather than at the output alone. DeltaLoss is not attracted to high prediction error alone.}
    \label{fig:noise_cifar_variants}
\end{figure}

\begin{figure}[ht]
    \begin{center}
        \includegraphics[width=0.95\linewidth]{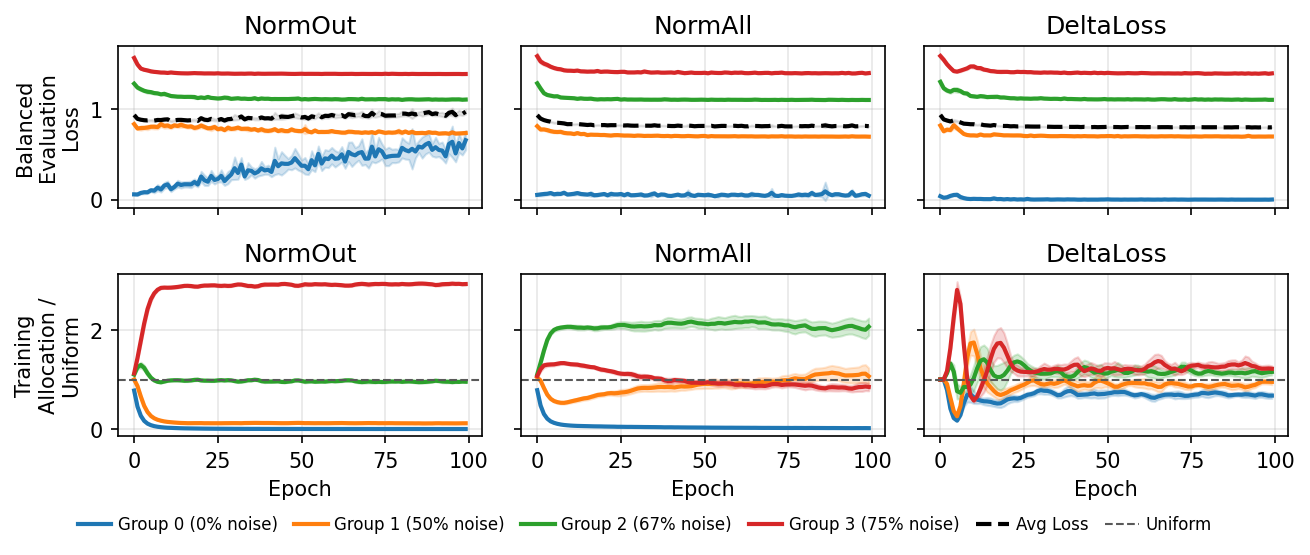}
    \end{center}
    \caption{MNIST Noise Condition for NormOut, NormAll, and DeltaLoss. \textbf{Top row:} Per-group Balanced Evaluation Loss. \textbf{Bottom row:} Mean per-class training allocation within each group. The behavior is similar to CIFAR-10.}
    \label{fig:noise_mnist_variants}
\end{figure}

\clearpage
\section{Related Work: Extended Discussion}\label{sec:app_related}

\subsection{Learning Progress and Intrinsic Motivation}

Schmidhuber's formal theory of creativity \citep{schmidhuber2010formal} proposes that agents should maximize ``improvements of the world model'', the rate of learning. 
Oudeyer and Kaplan \citet{oudeyer2007intrinsic} develop this in developmental robotics. 
However, ``learning progress'' has never been given an agreed definition, and what the literature provides are operationalizations rather than definitions.
Prior work measures progress through auxiliary task performance or prediction accuracy \citep{lopes2012exploration, bellemare2016unifying, pathak2019self}, or through prediction gain \citep{schmidhuber1991possibility} and compression progress \citep{schmidhuber2010formal}, all of which read progress off the model's output and inherit both high variance and the ambiguities described in Section~\ref{sec:intro}.

GMC operationalizes the same idea at the level of model parameters rather than through external performance metrics.

\subsection{Curiosity-Based Exploration}

Curiosity-driven exploration methods typically address sparse rewards by using model prediction error as a proxy for learning progress. 
Rather than explicitly estimating improvement over time, which is statistically challenging and often high variance, these approaches reward states or transitions where the agent's predictive models incur high loss, under the assumption that such regions offer greater potential for learning. 
ICM~\citep{pathak2017curiosity} operationalizes this idea by defining intrinsic reward as the prediction error of a learned forward dynamics model, while constraining the underlying representation through an auxiliary inverse-dynamics objective that encourages features to be informative about the agent's actions. 
RND \citep{burda2019exploration} uses prediction error against a fixed random target network as a novelty signal, while VIME \citep{houthooft2016vime} employs Bayesian information gain in the model's belief state. 
Despite their differences, these methods share a common foundation: instantaneous model loss is used as a surrogate for learning progress.

This surrogate introduces a fundamental limitation. 
High prediction error may arise either from learnable structure or from irreducible stochasticity, yet scalar loss provides no mechanism to distinguish between the two. 

\paragraph{ICM's Controllability Filter: Overfiltering and Underfiltering.}
In ICM, the inverse-dynamics objective is intended to mitigate this issue by filtering out uncontrollable sources of variation. 
However, this mechanism introduces two complementary failure modes.

First, overfiltering: by biasing representations toward information that is locally predictive of action, ICM disfavors state information whose relevance is temporally delayed or only indirectly connected to control. 
In environments where meaningful cues are revealed during phases of limited or no controllability, such as observations encountered during free fall after an irreversible action, these cues provide little or no inverse-dynamics learning signal. 
As a result, they are weakly represented or discarded, despite being critical for downstream task performance. 
This reflects a structural bias toward short-horizon controllability rather than long-term task relevance.

Second, underfiltering: the inverse-dynamics objective does not explicitly penalize retaining information unrelated to action prediction. 
In regions of the state space where actions have little influence on outcomes, the inverse loss becomes uninformative, providing no incentive to suppress irrelevant or stochastic features. 
Consequently, representations may continue to encode uncontrollable noise, and the forward-model prediction error, and thus intrinsic reward, remains persistently high. 
The agent is therefore still attracted to stochastic transitions, reproducing the noisy-TV failure mode the inverse model was intended to prevent.

Together, these issues show that while ICM reframes curiosity through controllability-aware representation learning, it neither guarantees the preservation of task-relevant but uncontrollable information nor reliably excludes irreducible randomness. 
The fundamental ambiguity of prediction error as a proxy for learning progress remains unresolved.

Recent work on multi-step inverse dynamics \citep{efroni2021provably, islam2023principled} addresses the underfiltering problem specifically: by conditioning on actions over multiple timesteps, these methods can provably separate endogenous (controllable) from exogenous (uncontrollable) state components, filtering distractors that single-step inverse dynamics misses.
However, multi-step conditioning requires a sufficiently long action horizon to capture the relevant dynamics, introducing practical challenges around tractability and variance that grow with horizon length.
Moreover, these methods improve the representation, not the measurement: the forward model's prediction error remains the intrinsic signal.

\paragraph{RND's One-Shot Geometry.}
RND defines intrinsic reward as prediction error to a fixed random target. 
While effective as a novelty heuristic, this design introduces two distinct limitations.

First, the induced representation geometry is arbitrary: because the target network is randomly initialized and never trained, the distances and novelty measures it induces bear no relationship to task-relevant structure. 
Random features may amplify or compress certain regions of the state space for reasons entirely unrelated to controllability, learnability, or reward. 
This can cause the agent to over-explore regions that merely happen to have high random-target prediction error, while under-exploring regions with more meaningful learning potential.

Second, RND's novelty signal is one-shot: because the reward decays monotonically as the predictor distills the fixed target, RND provides no mechanism to revisit regions that were encountered early in training but were not yet learnable with the agent's then-current representation capacity. 
If a region requires foundational skills or abstractions that have not yet been developed, the agent may visit it once, fail to learn, and then never return, even after acquiring the necessary prerequisites. 
This stands in contrast to learning-progress signals that can detect when previously-difficult regions become newly learnable.

\paragraph{Complementarity.}
ICM determines \emph{what features to learn} and RND \emph{what to treat as novel}, while GMC scores \emph{whether learning is occurring} on whichever representation is used, so one could replace prediction error with GMC in either architecture.

Recent work addresses other aspects of exploration. 
Never Give Up \citep{badia2020never} combines episodic and life-long novelty, Go-Explore \citep{ecoffet2019go} uses archive-based exploration, and counting-based methods \citep{bellemare2016unifying, ostrovski2017count} reward rarely-seen states. 
Goal-conditioned approaches \citep{andrychowicz2017hindsight, pong2019skew} learn to reach diverse goals. 
These focus on state-space coverage (discovering unknown unknowns), while GMC focuses on efficient learning within known unknowns. 
Combining coverage-based methods with learning-progress signals is a promising direction.

\subsection{Curriculum Learning}

Curriculum learning \citep{bengio2009curriculum} posits that learning is more efficient when examples progress from easy to hard. 
Automated approaches include teacher networks \citep{graves2017automated}, teacher-student frameworks \citep{matiisen2017teacher}, and self-paced learning \citep{kumar2010self}. 
Most require explicit difficulty rankings or auxiliary teacher models.

In our controlled partitions, GMC induces a prioritization that begins with the groups whose loss is falling fastest and defers the rest, reproduced across three different class-to-group assignments (Appendix~\ref{sec:app_repartition}).
Tasks where learning is rapid and persistent receive high scores. Tasks that are difficult (weakly aligned gradients), too noisy (incoherent gradients), or already learned (small gradients from saturation) receive low scores.
Curiosity, by contrast, prioritizes tasks by highest loss first, and both signals shift focus over time.
The observed ordering arises without manual tuning, but it is not a general guarantee of a convergence benefit, which remains task-dependent.

\subsection{Momentum in Optimization}

Classical momentum \citep{polyak1964some} accumulates gradients to accelerate learning and reduce oscillation. 
Modern variants include Nesterov acceleration \citep{nesterov1983method} and adaptive methods like Adam \citep{kingma2015adam}.

We repurpose momentum from an optimization tool to a learning signal. 
Rather than using momentum to smooth updates, we use it as a record of \emph{which parameters are currently moving}. 
High coupling (positive or negative) between a sample's gradient and momentum indicates the sample affects actively-changing parameters.
Low coupling indicates it affects stagnant parameters. 
This perspective aligns with work on optimization dynamics \citep{li2018visualizing} showing that successful optimization navigates with consistent patterns despite high dimensionality.

\subsection{Noise Robustness and Uncertainty}

Bayesian approaches \citep{gal2016dropout, kendall2017uncertainties} and ensemble methods \citep{osband2016deep, pathak2019self} provide mechanisms for estimating uncertainty, often useful for capturing epistemic uncertainty and, in some formulations, separating it from aleatoric noise.
However, they introduce additional computational overhead and require an additional scoring or decision rule to translate uncertainty into sample prioritization.

Our approach is more direct: random noise produces gradients pointing in inconsistent directions, which cancel in momentum over a sufficiently long averaging window.
Only consistent structure builds up. 
In our controlled setting, gradient-momentum coupling thereby filters noise while requiring no additional forward passes or ensemble maintenance.

\end{document}